%% file: main_arxiv.tex
\documentclass[10pt,twocolumn,letterpaper]{article}

\usepackage[pagenumbers]{cvpr} % To force page numbers, e.g. for an arXiv version

\input{preamble}

\newcommand{\method}{\textsc{SpotLight}\xspace}
\newcommand{\revise}[1]{{\color{blue}#1}}

\newcommand{\myparagraph}[1]{\noindent\textbf{#1}\:}

\newlength{\mywidth}

\definecolor{cvprblue}{rgb}{0.21,0.49,0.74}
\usepackage[pagebackref,breaklinks,colorlinks,citecolor=cvprblue]{hyperref}
\usepackage[capitalize]{cleveref}

\def\paperID{297} % *** Enter the Paper ID here
\def\confName{3DV\xspace}
\def\confYear{2026\xspace}

\title{\method: Shadow-Guided Object Relighting via Diffusion \vspace{-6pt}}

\author{Frédéric Fortier-Chouinard$^1$ \quad
Zitian Zhang$^1$ \quad
Louis-Etienne Messier$^1$
\\
Mathieu Garon$^2$ \quad
Anand Bhattad$^3$ \quad
Jean-Fran\c{c}ois Lalonde$^1$ \\
$^1$Université Laval, $^2$Depix Technologies, $^3$Toyota Technological Institute at Chicago \\
\normalsize{\texttt{\url{https://lvsn.github.io/spotlight}}}
\vspace{-6pt}
}

\begin{document}
\input{figs/figure_teaser_new}
\vspace{-6pt}
\maketitle

\input{sec/abstract}    
\input{sec/intro}
\input{sec/relwork}

\input{sec/methodology}

\input{sec/evaluation}
% \input{sec/2d_object_relighting}
\input{sec/extensions}
\input{sec/discussion}
\myparagraph{Acknowledgements.}
This research was supported by FRQNT scholarship 328810, NSERC grants RGPIN 2020-04799 and ALLRP 586543-23, Mitacs and Depix. Computing resources were provided by the Digital Research Alliance of Canada. The authors thank Zheng Zeng, Yannick Hold-Geoffroy and Justine Giroux for their help as well as all lab members for discussions and proofreading help.

{
    \small
    \bibliographystyle{ieeenat_fullname}
    \bibliography{main}
}

\end{document}

%% file: preamble.tex
\usepackage{graphicx}
\usepackage{amsmath}
\usepackage{amssymb}
\usepackage{booktabs}
\usepackage{mathtools}
\usepackage{array}
\usepackage{soul}
\newcolumntype{P}[1]{>{\centering\arraybackslash}p{#1}}
\usepackage{multirow}
\usepackage{float}      % For handling figure placement
\usepackage{afterpage}
\usepackage{makecell}
\usepackage{scalerel,xparse}
\usepackage{capt-of,etoolbox}
\usepackage{tikz}
\usepackage{algorithm}
\usepackage{algpseudocode}
\usepackage{wrapfig,lipsum,booktabs}

\usepackage{color}
\usepackage{colortbl}

\usepackage{dblfloatfix}

%% file: figs/figure_teaser_new.tex
\newcommand{\teaseroverlay}[1]{\llap{\raisebox{1.7cm}{\frame{\includegraphics[height=1.3cm]{#1}}}\hspace{1.5cm}}}
\newcommand{\teaseroverlayzoomtop}[1]{\llap{\raisebox{2cm}{\frame{\includegraphics[height=1.7cm,trim={2cm, 0cm, 2cm, 4cm},clip]{#1}}}\hspace{0cm}}}
\newcommand{\teaseroverlayzoom}[1]{\llap{\raisebox{1.7cm}{\frame{\includegraphics[height=1.3cm,trim={4cm, 2cm, 4cm, 6cm},clip]{#1}}}\hspace{1.5cm}}}
\newcommand{\teaseroverlaypad}{\llap{\raisebox{1.7cm}{}\hspace{1.5cm}}}

\twocolumn[{%
\renewcommand\twocolumn[1][]{#1}%
\maketitle
\begin{center}
    \centering
    \footnotesize
    \setlength{\tabcolsep}{2pt}
    \setlength{\mywidth}{0.185\linewidth}
    \captionsetup{type=figure}
    \begin{tabular}{cccccc}
    \includegraphics[width=\mywidth]{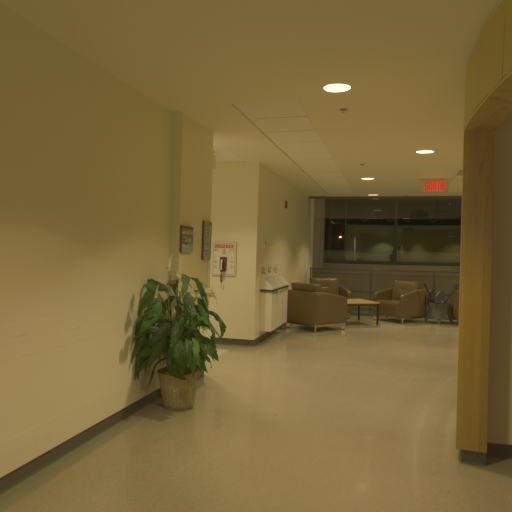} & 
    \includegraphics[width=\mywidth]{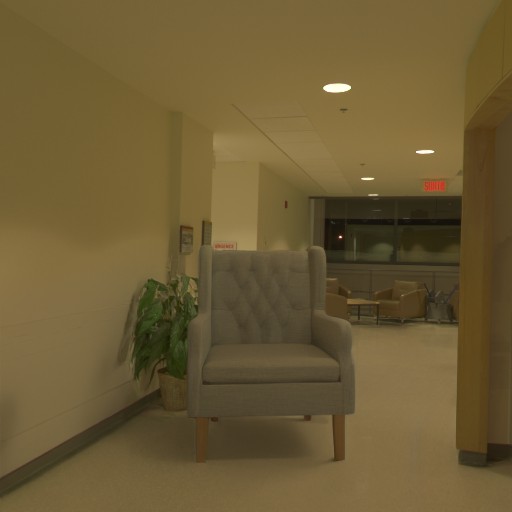} & ~~~ &
    \includegraphics[width=\mywidth]{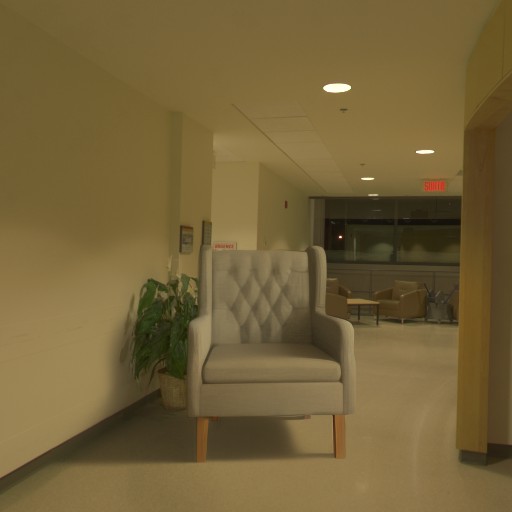} 
    \teaseroverlayzoomtop{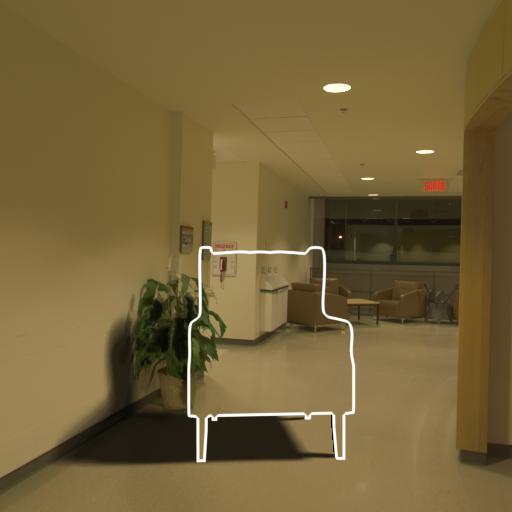} 
    &
    \includegraphics[width=\mywidth]{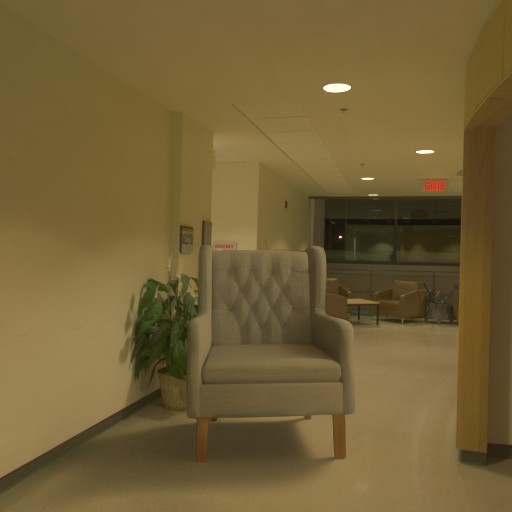}
    \teaseroverlayzoomtop{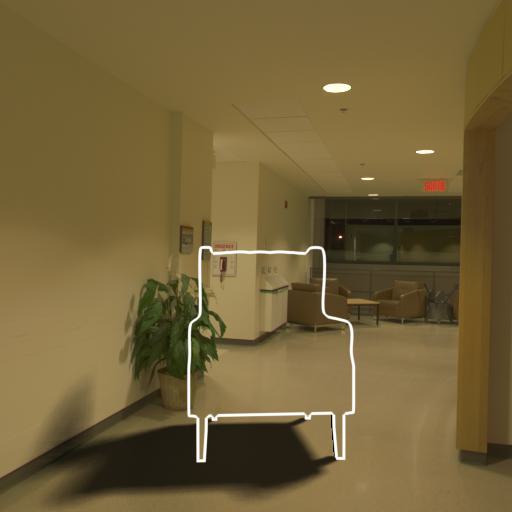} 
    &
    \includegraphics[width=\mywidth]{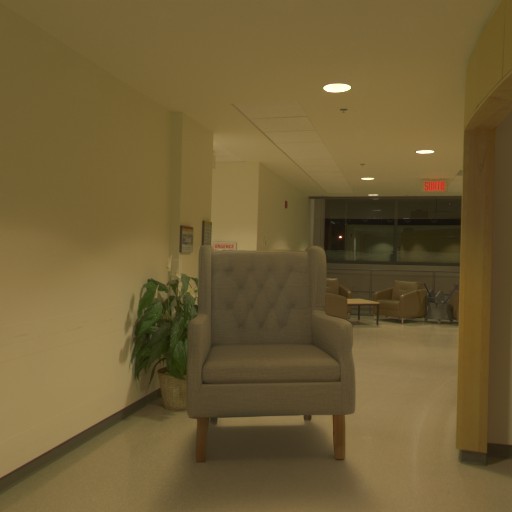}
    \teaseroverlayzoomtop{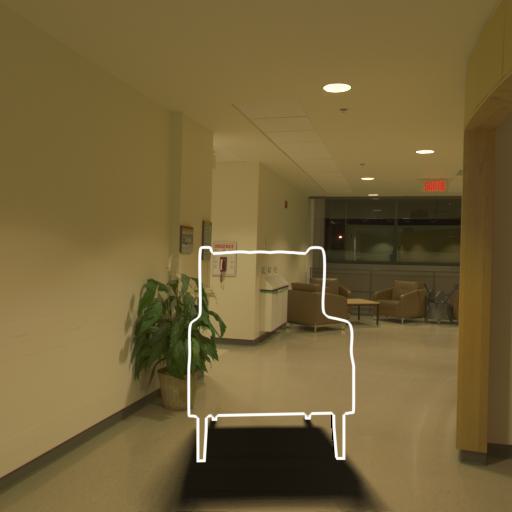} \\
    % &
    % \includegraphics[width=\mywidth]{figs/teaser/shadow_comp_v3_w_latent_edge_boost/9C4A2857-dc0a9513b2_04_crop_B07QGSZ4LN/default/0005.jpg}
    % \teaseroverlayzoomtop{figs/teaser/shadow_comp_v3_w_latent_edge_boost/9C4A2857-dc0a9513b2_04_crop_B07QGSZ4LN/intermediate/shadow_positive_outline_0005.jpg} 
    % &
    % \includegraphics[width=\mywidth]{figs/teaser/shadow_comp_v3_w_latent_edge_boost/9C4A2857-dc0a9513b2_04_crop_B07QGSZ4LN/default/0004.jpg}
    % \teaseroverlayzoomtop{figs/teaser/shadow_comp_v3_w_latent_edge_boost/9C4A2857-dc0a9513b2_04_crop_B07QGSZ4LN/intermediate/shadow_positive_outline_0004.jpg} 
    Background image & Previous method & & \multicolumn{3}{c}{\method enables shadow-guided lighting control}
    \end{tabular}
    %\captionof{figure}{\method generates realistic local lighting by simply moving a shadow around. Our method generates realistic reflections and shadows while only requiring simple edits to the latents of a pre-trained diffusion-based neural renderer. Here, the virtual light source is at an elevation of 45 degrees and rotates in azimuth from from right to left in camera-space.}
    \vspace{-5pt}
    \captionof{figure}{When inserting a piece of virtual furniture (chair) in an image, (left) existing diffusion-based renderers produce static composites without lighting control. In contrast, (right) \method enables \emph{shadow}-guided lighting control: a user specifies the desired shadows of the inserted object (inset)---\method realistically blends the shadows and relights the object appropriately, acting as a virtual spotlight for the inserted object. Through our proposed shadow conditioning, we show that existing diffusion-based renderers can be guided to achieve realistic, controllable relighting without requiring any additional training.
    }
    \label{fig:teaser} 
    \vspace{-5pt}
\end{center}
}]

%% file: sec/abstract.tex
\begin{abstract}
Recent work has shown that diffusion models can serve as powerful neural rendering engines that can be leveraged for inserting virtual objects into images. However, unlike typical physics-based renderers, these neural rendering engines are limited by the lack of manual control over the lighting, which is often essential for improving or personalizing the desired image outcome. In this paper, we show that precise and controllable lighting can be achieved without any additional training, simply by supplying a coarse \emph{shadow} hint for the object. Indeed, we show that injecting only the desired shadow of the object into a pre-trained diffusion-based neural renderer enables it to accurately shade the object according to the desired light position, while properly harmonizing the object (and its shadow) within the target background image. 
Our method, \method, is entirely training-free and leverages existing neural rendering approaches to achieve controllable relighting.
We show that \method achieves superior object compositing results, both quantitatively and perceptually, as confirmed by a user study, outperforming existing diffusion-based models specifically designed for relighting. We also demonstrate other applications, such as hand-scribbling shadows and full-image relighting, demonstrating its versatility.
% Our framework, \method, injects the shadow information in pre-trained, diffusion-based neural renderers and 
% We present \method, a framework for local lighting control over an image. \todo{emphasize that our goal is to use shadows as a knob on the lighting, and it's pretty cool that it works} By only rendering coarse cast shadows of an object, or even hand-drawn shadows, \method can realistically relight the object, making its shading consistent with both the desired shadow direction and the background, while refining the shadow. Our method was applied to two diffusion-based neural renderers found in the literature and didn't require any training to enable control over the lighting. We further compare our method against existing diffusion-based lighting control models and show superior results quantitatively and through a user study. Finally we propose a novel dataset for evaluating local lighting control methods.
\end{abstract}

%% file: sec/intro.tex
\section{Introduction}
\label{sec:intro}
% \todo{compare changes with previous versions --- do they address reviewers' issues or make things clearer?}

Object relighting---the process of inserting a virtual object in an image and lighting it appropriately---has traditionally been confined to 3D graphics pipelines, where lighting is explicitly modeled and rendered. In contrast, object relighting in real photographs remains a fundamentally ill-posed problem since it requires inferring the physical properties of the scene: its geometry, materials, and illumination are not observed, and their complex interactions---such as specular reflections, soft shadows, or material-dependent reflectance---must be inferred or approximated. 
% These challenges become particularly acute when attempting to realistically insert or relight objects within existing photographs.

%Image relighting---the process of modifying the lighting conditions in a scene---is a capability that has long been confined solely to the field of computer graphics, where a user is free to control the lighting conditions in the 3D scene before launching the rendering process. When one is given a photograph, however, no such control is possible. Relighting then becomes a hard task because of the complex  interactions between light, material properties, and surface characteristics, where different materials respond distinctly to lighting variations through specular highlights, surface orientation, and texture. The recent rise of powerful image generation models (\eg, \cite{rombach2022high}) has enabled exciting possibilities for image relighting~\cite{kocsis2024lightit,iclight}.

Recent advances in image generative models, especially diffusion-based methods~\cite{rombach2022high}, have opened up new possibilities for object relighting~\cite{zhang2025iclight,zhang2025zerocomp,liang2025diffusionrenderer}. These models encode strong priors over the appearance of scenes under varied lighting, allowing implicit reasoning about illumination, material, and geometry. Yet, precise and localized relighting control remains limited, particularly in the context of virtual object insertion, where a single object must be relit.

We focus on the problem of local lighting control for object compositing using generative models. Specifically, we aim to relight an inserted object such that the object conforms to a specified lighting direction and intensity around it, while blending it naturally with the background. Achieving such control is challenging: the relighting must align with physical cues like shadows and shading, yet remain flexible enough to work across varied materials, perspectives, and image contexts. Diffusion models offer powerful priors, but they are not inherently controllable without carefully designed guidance.

We present \method, a training-free method for plausible, realistic, and controllable object relighting by leveraging the rich prior knowledge encoded in pre-trained, diffusion-based neural renderer. Rather than obtaining specialized high-quality data and training a model specifically for the task of object relighting (e.g., \cite{zeng2024dilightnet,jin2024neural_gaffer}), our key insight is to drive the relighting process by supplying a coarse representation of the desired \emph{shadow} of the object to a general-purpose neural renderer. Not only is a 2D shadow an intuitive input for human control, but we also find that diffusion models can effectively utilize it. We call our method \method because it guides the diffusion process in a manner analogous to a physical spotlight, selectively illuminating the object to insert in the image.

%Our core technical contribution is this shadow guidance approach that refines the input shadow while realistically relighting the virtual object. 
\method exploits classifier-free guidance within existing pre-trained diffusion-based renderers and progressively harmonizes imperfect shadows with the background while relighting the object. Our guidance mechanism steers the rendering process and modifies the appearance of the inserted composite, resulting in local lighting control. This mechanism is applied to recent diffusion models pre-trained to render images from input intrinsic maps (depth, normals, albedo, etc.)~\cite{zhang2025zerocomp,zeng2024rgb}.

% This provides a straightforward means of influencing the final appearance of the object.
% Our key insight is to control object lighting by supplying a coarse representation of the desired \emph{shadow}, offering a straightforward means of influencing the final appearance of the composition. 

% \Cref{fig:teaser} shows an example of our approach: a synthetic shadow, rendered using standard shadow mapping~\cite{eisemann2011realtime}, is provided as guidance for the desired lighting direction. \method harmonizes this shadow into the background and relights the inserted object accordingly. Our method builds on classifier-free guidance in diffusion models to progressively refine the relit object, blending realistic lighting effects with minimal artifacts and no additional training.}

\Cref{fig:teaser} shows an example of such shadow-based lighting control: the shadow of the chair is rendered using basic shadow mapping~\cite{eisemann2011realtime} according to the desired lighting direction. From this, \method realistically blends the input guiding shadow within the background, and relights the virtual object to match the desired lighting. % at each denoising step. % This process requires no additional training.  % \todo{Justine: switch order of 1st and 2nd sentence} %\todo{CFG isn't the part that harmonizes the shadows, this should be clarified} 

In summary, our contributions are as follows. First, we introduce a framework that enhances pre-trained diffusion renderers with object relighting control without requiring additional training. We demonstrate superior quality, both perceptually through user studies and quantitatively, compared to competitive specialized models. Finally, we release an evaluation dataset tailored for lighting control in object insertion, which will be made available upon publication.
% 4) \todo{fix this}? \todo{Not sure if that's a contribution: 4) we perform a thorough quantitative and qualitative evaluation of our method and SOTA diffusion-based lighting control methods.}

% \todo{Copied from sec. 3: We propose a new method of injecting coarse shadows into a diffusion-based neural renderers, without requiring any additional training. We focus only on shadows since they indirectly provide a precise image-based cue on the 3D lights directions. We assume that the shadows fed to our method will be imperfect and contain visible artifacts caused either by the use of shadow mapping, which doesn't fully model the light interactions, the use of imperfect, estimated geometry and the simple alpha compositing of shadows, which ignores possible shadow overlap. We thus want our method to be robust to those artifacts, by harmonizing them with the background. We propose to use the neural renderer's priors on realistic images to progressively harmonize, at each denoising step, the desired shadow with the noisy latents.}

%% file: sec/relwork.tex
\section{Related work}
\label{sec:relwork}
% 1. Diffusion models for image generation; 2. Finetuning UNet or Using Controlnet to add more control to the generation; 3. Image generation manupulations by changing the latents. 4. Classifier free guidance;

\myparagraph{Conditional image generation.} Diffusion-based generative models have become the de facto choice for image generation due to their ability to produce high-quality images~\cite{sohl2015deep, ho2020denoising,song2020denoising, rombach2022high}. These models offer control mechanisms for various modalities, such as text~\cite{ramesh2022hierarchical,rombach2022high}, image content~\cite{saharia2022palette, gal2022image}, intrinsic images~\cite{zhang2025zerocomp,luo2024intrinsicdiffusion,zeng2024rgb,liang2025diffusionrenderer}, classifier guidance~\cite{dhariwal2021diffusion} and cross-modal data~\cite{brooks2023instructpix2pix, ye2023ip}. Notably, training-free guidance methods~\cite{meng2022sdedit, epstein2023diffusion, avrahami2022blended, avrahami2023blendedlatent,ho2021classifierfree} have enabled image editing capabilities without specific training, forming the basis of our approach.

\myparagraph{Image relighting} modifies the global or local shading of an image without changing other properties such as geometry and materials. Previous methods 
%for style transfer~\cite{gatys2016image, luan2017deep} 
for harmonization~\cite{reinhard2001color, pitie2005n, lalonde2007using, tsai2017deep} do not prioritize physically accurate shading of the objects. Recently, \cite{poirier2024diffusion} utilized a single-view multi-illumination dataset~\cite{murmann2019multi} to enable direct control over the dominant lighting direction. Retinex theory has been leveraged~\cite{xing2024retinex,bhattad2024stylitgan} for relighting indoor scenes. IC-Light~\cite{zhang2025iclight} enforces consistency in appearance 
to relight portraits and various objects. LumiNet~\cite{xing2025luminet} and ScribbleLight~\cite{choi2025scribblelight} achieve image relighting through illumination transfer or user-defined scribbles, respectively.
%but as our results show does not generalize to complex objects like indoor furniture \todo{I think the reason is that it does not enforce consistency with the rest of the scene, and has limited physical control}. 
%allowing for the blending of multiple light sources to achieve precise relighting effects. 
Some methods specialize in outdoor scenes, using geometric priors like depth~\cite{griffiths2022outcast} or normalized coordinates~\cite{kocsis2024lightit} or both~\cite{lin2025urbanir}. These methods however cannot handle lighting on specific region or objects in the scene.  
Recent work on object relighting has evolved from consistency-based approaches in intrinsic images~\cite{bhattad2022cut} to harmonization of foreground-background albedo~\cite{careaga2023intrinsic}. 
% Bhattad \etal~\cite{bhattad2022cut} relit inserted objects via enforcing the consistency on intrinsic images, which is later revisited by Careaga et al~\cite{careaga2023intrinsic} by harmonizing the foreground albedo with the background. 
% ZeroComp~\cite{zhang2025zerocomp} learns to composite virtual objects in a zero-shot fashion, by treating diffusion models as neural renderers, but does not offer lighting control. % RGB$\leftrightarrow$X~\cite{zeng2024rgb} also enables zero-shot compositing, with limited, text-guided control lighting control.
% DiffusionRenderer~\cite{liang2025diffusionrenderer} requires a long and approximate lighting estimation process to annotate each image in the training set.
Recently, diffusion models have been adapted for the task of rendering from partial intrinsic maps~\cite{zhang2025zerocomp,zeng2024rgb,liang2025diffusionrenderer}, which can be leveraged to achieve zero-shot object composition. However, these methods have limited lighting control. ZeroComp~\cite{zhang2025zerocomp} provides no lighting control, RGB$\leftrightarrow$X~\cite{zeng2024rgb} can only provide lighting control through a text prompt, and DiffusionRenderer~\cite{liang2025diffusionrenderer} requires a prohibitively long and approximate lighting estimation process to annotate each sample in the training set.
Other diffusion-based approaches, such as Neural Gaffer~\cite{jin2024neural_gaffer}, DiLightNet~\cite{zeng2024dilightnet} and IllumiNeRF~\cite{zhao2024illuminerf} utilize HDR environment maps and multi-lighting renders, but are not designed for object composition in existing images. LightLab~\cite{magar2025lightlab} achieves fine-grained control over existing light sources, but does not handle object insertion. Careaga et~al.~\cite{careagaRelighting} propose a sophisticated inverse rendering method for data generation, wheras \method is training-free. IntrinsicEdit~\cite{lyu2025intrinsicedit} proposes a training-free method for lighting-based editing, but, unlike our approach, lacks explicit control over object-specific virtual lighting effects. We bridge this gap by introducing a shadow-guided strategy for object relighting that leverages existing models without any additional training. 
%RGB$\leftrightarrow$X \cite{zeng2024rgb} decomposes images into intrinsic maps, then composites and synthesizes them using material- and lighting-aware diffusion models, which allows text prompts for light conditioning. Neural Gaffer~\cite{jin2024neural_gaffer} and DiLightNet~\cite{zeng2024dilightnet} are lighting controlled diffusion models that leverage either HDR environment maps, or the renders under different lightings. While effective, these methods lack control over object relighting. We propose a controllable neural renderer for local relighting guided by a shadow.

% 2. Local relighting:
% Intrinsics-based neural rendered: ZeroComp~\cite{zhang2025zerocomp}, RGB$\leftrightarrow$X~\cite{zeng2024rgb}, Intrinsic ControlNet~\cite{anonymous2024intrinsiccontrolnet}. Alchemist~\cite{sharma2024alchemist} only for global object material edits
% Relight object, without any effect by the bg: DiLightNet~\cite{zeng2024dilightnet} and Neural Gaffer~\cite{jin2024neural_gaffer}
% Facial relighting: DifFRelight, DiFaReli, COMPOSE \cite{hou2024composecomprehensiveportraitshadow}

\myparagraph{Explicit lighting estimation} approaches infer HDR lighting from a single image~\revise{\cite{gardner2017sigasia, garon2019cvpr, li2020cvpr, zhu2022irisformer,phongthawee2024diffusionlight,liang2024photorealistic}}, though they generally lack controllability. Parametric models~\cite{gardner2019iccv, weber2022editable, dastjerdi2023everlight} or GAN inversion methods~\cite{wang2022stylelight} offer more control, but these methods require physics-based rendering engines for generating the composite image---here, we focus instead on neural renderers.

\myparagraph{Shadow generation.} Generative models for rendering realistic shadows~\cite{zhang2019shadowgan, liu2020arshadowgan,hong2022shadow,liu2024shadow}, or harmonizing rendered shadows with the image~\cite{valenca2023shadow} have been proposed. Other methods enable controllable shadows for 3D~\cite{sheng2021ssn} and 2D objects~\cite{sheng2022controllable, sheng2023pixht}. Our approach can leverage existing shadow generation methods (e.g., \cite{tasar2024controllable}) and shows that diffusion-based neural renderers can be conditioned on such shadows to achieve object relighting.

\myparagraph{Image intrinsics.} Decomposing images into albedo and shading has long been studied~\cite{barrow1978recovering, kovacs2017shading, philip2019multi, wu2023measured, careaga2023intrinsic}. Numerous works have also focused on estimating scene geometry, for example, depth~\cite{ranftl2020towards, bhat2023zoedepth, yin2023metric3d, ke2024repurposing, yang2024depth, depth_anything_v2} and normals~\cite{eftekhar2021omnidata, bae2021estimating, ranftl2021vision, bae2024rethinking, ye2024stablenormal}. Recently, approaches such as \cite{li2020cvpr,zhu2022irisformer} jointly infer different intrinsic maps including shape, spatially-varying lighting, scene geometries, and materials. Conversely, generative models have shown a powerful internal understanding of intrinsic properties 
%with minimal to no fine-tuning
~\cite{bhattad2024stylegan, du2023generative, ke2024repurposing, ye2024stablenormal, anonymous2024intrinsiccontrolnet, luo2024intrinsicdiffusion}. We similarly leverage diffusion models' ability to interpret intrinsic maps to generate realistic images as a foundation for rendering image composites.

%% file: sec/methodology.tex
%!TEX root = ../main.tex
\input{figs/figure_pipeline}

\input{figs/figure_qualitative_control_3dv}

\section{Background: diffusion renderers}
\label{sec:background-diffusion-renderers}

Recent work has shown that diffusion models can be adapted to generate photorealistic images from intrinsic maps including materials (e.g., albedo, roughness, metalness), geometry (e.g., surface normals and depth), and shading. This capability was built using a ControlNet in ZeroComp~\cite{zhang2025zerocomp} or full finetuning in RGB$\leftrightarrow$X~\cite{zeng2024rgb}. We refer to this new class of conditional generation models as ``diffusion renderers''. 
% For instance, ZeroComp~\cite{zhang2025zerocomp} is able to generate realistic images even when a large portion of the shading map is masked out, therefore inferring local shadows and shading effects, while maintaining consistency with the scene. The training data for such methods can come from synthetic scenes or annotating real images using off-the-shelf intrinsic decomposition methods.

Specifically, ZeroComp~\cite{zhang2025zerocomp}, which we will use in \cref{sec:evaluation}, is trained to generate images from albedo, normals, depth, and partially-masked shading. It can be used for compositing in a zero-shot manner by alpha-compositing the intrinsics of the background and the object, and by masking out a region on and around the object in the shading map. The network then generates the full image with realistic shading. The intrinsics of the object are rendered using simple graphics shaders, while those of the background are estimated using pre-trained networks (e.g., \cite{bhat2023zoedepth} for depth, see \cref{sec:eval-specifics}). In the case of RGB$\leftrightarrow$X~\cite{zeng2024rgb}, their X$\rightarrow$RGB network achieves object insertion by accepting as input alpha-composited albedo, normals, and a masked image. They then train an inpainting version of their network to generate the final image. Unlike traditional rendering, these methods offer no control over the lighting conditions.

In this work, we use such a pre-trained diffusion renderer and show that, by incorporating an \emph{approximate} guiding shadow, we can steer the diffusion model to enable local lighting control, \emph{without any additional training}. 
% analogous to a spotlight moving around the object. 
% We dub our approach \method, which provides a simple and intuitive means to control local lighting. Our setup retains the same input configuration as conventional diffusion rendering, with the sole addition of a shadow. 
We now describe our approach in detail.

\section{\method}
\label{sec:methodology}

% Our goal is to adapt a conditional diffusion model trained to render a photorealistic image based solely on 2D intrinsic maps (\eg albedo, normals, depth, partial shading) to be able to control the lighting on objects in the scene. Since all of these methods work in the 2D image space, we need to define a novel and simple way to provide a full description of the light sources in image space. We therefore use the shadow cast by the object as the lighting conditioning.

% Given a pre-trained diffusion-based neural renderer, 

%The diffusion-based neural renderers presented in \cref{sec:background-diffusion-renderers} are currently limited by their inability to directly control lighting \todo{for fine-grained object composting}. \method addresses this limitation by enabling precise control over the lighting of the object. \Cref{fig:pipeline} presents an overview of the method, with each component described in detail below. After describing the required inputs, we present the two main steps of our approach, namely shadow blending and object relighting.

%Diffusion-based neural renderers inherently lack explicit control over lighting. 
We employ a user-provided, guiding shadow to steer a pre-trained diffusion renderer (c.f. \cref{sec:background-diffusion-renderers}) to follow the desired lighting direction. 
% thereby enforcing matching illumination on the inserted object. 
Our key insight is that even an approximate shadow encapsulates enough important lighting information and, by integrating this cue in the denoising process, we can simultaneously refine the composite appearance and its shadow, yielding a natural blend with the scene.
% This cue effectively defines a virtual light source around the composite, analogous to a spotlight. 

\subsection{Approach overview}
\label{sec:user-input}

An overview of \method is illustrated in \cref{fig:pipeline}. In addition to the intrinsic maps for the backbone diffusion renderer, our approach requires two inputs: an object mask $\mathbf{m}_\text{obj}$ and a user-specified guiding shadow $\mathbf{m}_\text{shw}$ which guides the latent diffusion process. The guiding shadow can be obtained in several ways---in this paper, we experiment with fast rasterization~\cite{eisemann2011realtime}, a trained shadow generation model~\cite{tasar2024controllable}, and hand-drawn scribbles. Without retraining the diffusion renderer, we steer its generative process by incorporating the guiding shadow into the latent space. % This not only adjusts the object's appearance but also progressively refines the shadow to better match the background illumination.

\subsection{Blending shadows with the background}

% \subsection{Blending shadows \revise{with the background}}
% % \subsection{Shadow-guided latent fusion}
% To fuse the guiding shadow with the latent representation, we update the noisy latent code $\mathbf{z}_t$ at each diffusion timestep, similar to~\cite{avrahami2023blendedlatent}. In our framework, the updated latent 
% \begin{equation}
%     \mathbf{\tilde{z}}_t = (1 - \beta \mathbf{m}_{\text{shw},\downarrow}) \odot \mathbf{z}_t + (\beta \mathbf{m}_{\text{shw},\downarrow}) \odot \text{noise}(\mathcal{E}(\mathbf{g}), t)
%     \label{eq:shadow-blending}
% \end{equation}
% serves as a soft constraint that steers the process toward a shadow-consistent output. Here, $\mathbf{g}$ \revise{is a rough} composite of the object albedo, the target shadow, and the background, \revise{which is encoded using the diffusion model's VAE encoder, $\mathcal{E}(\cdot)$. The parameter} $\beta$ controls the strength of the guiding shadow\revise{, which we set to a default value of 0.05}. The latent shadow mask $\mathbf{m}_{\text{shw},\downarrow}$ is obtained by downsampling the target shadow $\mathbf{m}_{\text{shw}}$ to the latent resolution \revise{(64$\times$64), with the edges dilated and increased in order to maintain the desired softness} (see supp. for details). \revise{Different from~\cite{avrahami2023blendedlatent}, we do not use the slow test-time decoder optimization nor the progressive mask shrinking steps.}

%Specific implementation details (e.g., mask dilation and resolution matching) are provided in the supplementary material.

Provided the guiding shadow 
$\mathbf{m}_{\text{shw}}$, we first need to ensure that the diffusion model will properly blend the shadow with the background. We take inspiration from Blended Latent Diffusion~\cite{avrahami2023blendedlatent} to progressively blend the noisy latents $\mathbf{z}_t$ at time $t$ with a noised VAE-encoded image which contains the shadow roughly composited over the image, $\mathbf{g}$. In practice, we obtain $\mathbf{g}$ by compositing the object albedo and the guiding shadow on the background. 
%$\mathbf{g} = \mathbf{x} \odot (1-\mathbf{m}_\text{shw})$.
The update made at each timestep to incorporate the desired shadow latents is
\begin{equation}
    \mathbf{\tilde{z}}_t = (1 - \beta \mathbf{m}_{\text{shw},\downarrow}) \odot \mathbf{z}_t + (\beta \mathbf{m}_{\text{shw},\downarrow}) \odot \text{noise}(\mathcal{E}(\mathbf{g}), t) \,,
    \label{eq:shadow-blending}
\end{equation}
where $\beta=0.05$ is the shadow latent weight and $\mathbf{m}_{\text{shw},\downarrow}$ is the shadow mask, bilinearly downsampled to the latent resolution ($64\times64$), with the edge opacity increased and dilated to maintain the desired shadow softness (see supp. for details). Noise is added to the encoded composite using the DDIM~\cite{song2020denoising} noise scheduler. Different from \cite{avrahami2023blendedlatent}, we do not use the slow test-time decoder optimization nor the progressive mask shrinking steps.

\subsection{Enhancing lighting control}
\label{sec:method-enhancing}

Simply running a diffusion renderer conditioned on the desired shadow direction yields perceptible, but subtle visual changes in the generated object (as will be shown in \cref{sec:ablations}). We propose to amplify the effect by running two parallel branches of the model: one which receives the shadow in the desired direction (positive branch), while the other has the shadow in the opposite direction (negative branch), obtained by shifting the light azimuth angle by $180^\circ$. Although the opposite light direction works best for the negative branch, we found that casting no shadow also provides a good negative sample.
%\todo{expand a bit: mention that there are several options for the negative branch. this and no-shadow, for example. makes little difference.}
From the outputs of the diffusion renderer (typically, $v$-predictions~\cite{salimans2022progressive}, i.e., linear combination of noise and estimated image), we then use classifier-free guidance to amplify the effect of the positive shadow on the lighting on the object:
\begin{equation} 
\begin{split}
\label{eq:modified_cfg}
    \mathbf{\tilde{v}}_t &= (1-\mathbf{m}_{\text{obj},\downarrow}) \odot \mathbf{v}_{t,\text{pos}} \\
    & + \mathbf{m}_{\text{obj},\downarrow} \odot ( \mathbf{v}_{t,\text{neg}} + \gamma( \mathbf{v}_{t,\text{pos}} - \mathbf{v}_{t,\text{neg}} )) \,,
\end{split}
\end{equation}
where $\gamma=3.0$ is the guidance scale, $\mathbf{m}_{\text{obj},\downarrow}$ is the object mask bilinearly downsampled to the latent resolution ($64\times64$). After  \cref{eq:modified_cfg}, we run a regular DDIM sampling step to obtain the latents $\mathbf{z}_{t-1}$.

% \subsection{Forcing object relighting}
% \subsection{Shading the object via dual-branch guidance}
% \label{sec:method-enhancing}
% 
% A key element of our approach is to amplify the impact of the guiding shadow on the object shading through classifier-free guidance (CFG)~\cite{ho2021classifierfree}. We process two parallel diffusion branches: a \emph{positive branch} that incorporates the desired shadow direction, and a \emph{negative branch} that suppresses an undesired shadow direction (\eg, via an opposing light direction). \Cref{fig:pipeline} shows our dual-branch guidance mechanism.  Classifier-free guidance then combines these branches:
% \begin{equation}
% \begin{split}
%     \mathbf{\tilde{v}}_t &= (1-\mathbf{m}_{\text{obj},\downarrow}) \odot \mathbf{v}_{t,\text{pos}} \\
%     &\quad+ \mathbf{m}_{\text{obj},\downarrow} \odot \Bigl( \mathbf{v}_{t,\text{neg}} + \gamma\bigl( \mathbf{v}_{t,\text{pos}} - \mathbf{v}_{t,\text{neg}} \bigr) \Bigr),
%     \label{eq:modified_cfg}
% \end{split}
% \end{equation}
% where $\gamma$ sets the guidance scale and $\mathbf{m}_{\text{obj},\downarrow}$ is the object mask downsampled to the latent resolution. This dual-branch strategy reinforces the intended lighting on the object while naturally blending it with the background.

\subsection{Final image synthesis}
After the guided diffusion process, the resulting latents $\mathbf{z}_{0}$ are decoded with the VAE decoder. The background preservation strategy of \cite{zhang2025zerocomp} is applied to ensure that only the object and its shadow are modified, yielding a composite image with natural local lighting.

%% file: figs/figure_pipeline.tex
\begin{figure*}[ht!]
    \centering
    \includegraphics[width=1.0\linewidth]{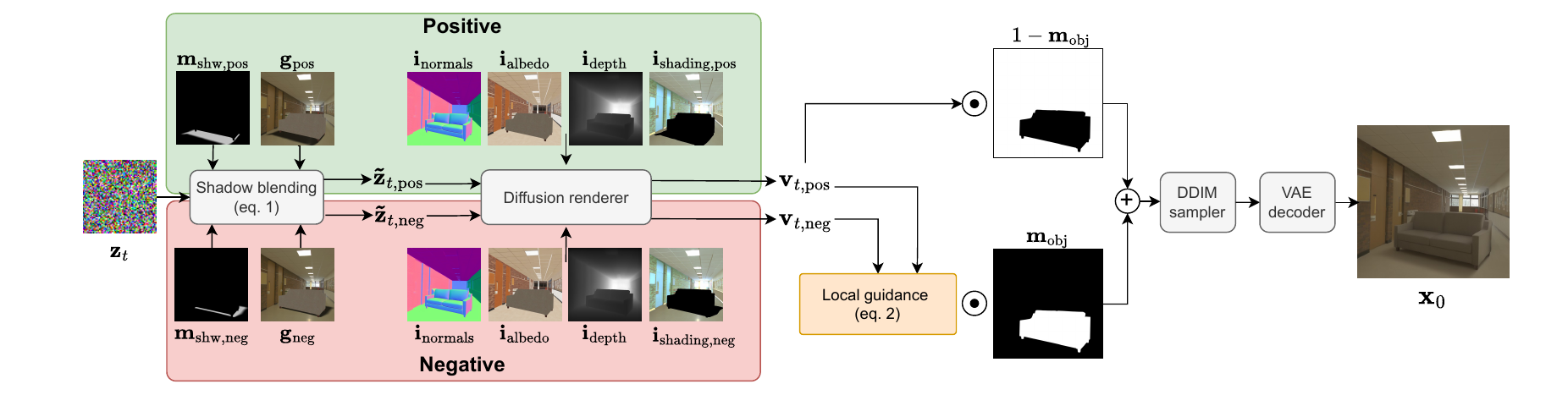}
    \vspace{-15pt}
    \caption{Overview of \method. We leverage diffusion-based neural renderers, pre-trained to render images from input intrinsics (normals, albedo, depth, and partial shading). Our proposed framework consists of two parallel branches: a positive branch (green), where the user-provided guiding shadow aligns with the desired light direction, and a negative branch (red), where the shadow is aligned with an opposite light direction. Initially, a latent blending operation merges the noisy latents with a rough scene composition using the background image, object albedo, and a coarse shadow. Both branches are then processed by a diffusion renderer conditioned on intrinsic images. Finally, relighting of the object is amplified using local guidance. 
    % \todo{add top part to figure, showing ``traditional rendering pipeline'' with diffusion renderers, for noobs (just an idea, maybe bad)---JF: I don't think that's needed}
    }
    \label{fig:pipeline}
\end{figure*}
% LINK to edit in draw.io:

% Version 2 (LATEST):
% https://drive.google.com/file/d/1OXfZJ8aFzuQWFsINWu2YaxKFqhJFWRIf/view?usp=sharing

% Version 1 (old but has all images):
% https://drive.google.com/file/d/1dJvyu4h98V23eAFxpH-IUfoavtNoFGls/view?usp=sharing

%% file: figs/figure_qualitative_control_3dv.tex
% \begin{strip}
%     \centering
%     \scriptsize
%     \setlength{\mywidth}{0.140\linewidth}
%     \def\arraystretch{0.25}
%     \setlength{\tabcolsep}{0.25pt}
%     \newcommand{\qualoverlay}[1]{\rlap{\raisebox{1cm}{\frame{\includegraphics[height=1.5cm]{#1}}}\hspace{0cm}}}
%     
%     \input{figs/qual_many_directions_iccv4/results_figure}
%     \captionof{figure}{Qualitative comparison against the baselines on our evaluation dataset. For each scene, we show the dominant shadow direction from the ``reference-based'' dataset, where the ground truth is available, and one from the ``user-controlled'' dataset. Observe how \method generates results that are visually closest to the ground truth (odd rows) and accurately match the guiding shadow (even rows) without changing the color of the object (IC-Light, DiLightNet), preserving the shape of the guiding shadow (ZeroComp+SDEdit), and generates more visible shading effects on the object (Neural Gaffer), all without being trained specifically for object relighting (DiLightNet, Neural Gaffer). Please zoom in and consult the supp. for additional qualitative results.
%         }
%     \label{fig:qual-usercontrol}
% \end{strip}

\begin{figure*}
    \centering
    \footnotesize
    \setlength{\mywidth}{0.140\linewidth}
    \def\arraystretch{0.25}
    \setlength{\tabcolsep}{0.25pt}
    \newcommand{\qualoverlay}[1]{\rlap{\raisebox{1cm}{\frame{\includegraphics[height=1.5cm]{#1}}}\hspace{0cm}}}
    
    \input{figs/qual_many_directions/results_figure}
    \caption{Qualitative comparison against the baselines on our evaluation dataset. For each scene, we show the dominant shadow direction from the ``reference-based'' dataset, where the ground truth is available, and one from the ``user-controlled'' dataset. Observe how \method generates results that are visually closest to the ground truth (odd rows) and accurately match the guiding shadow (even rows) without changing the color of the object (IC-Light, DiLightNet), preserving the shape of the guiding shadow (ZeroComp+SDEdit), and generates more visible shading effects on the object (Neural Gaffer), all without being trained specifically for object relighting (DiLightNet, Neural Gaffer). Please zoom in and consult the supp. for additional qualitative results.
        }
    \label{fig:qual-usercontrol}
\end{figure*}

%% file: figs/qual_many_directions/results_figure.tex
\begin{tabular}{ccccccc} 
\includegraphics[width=\mywidth]{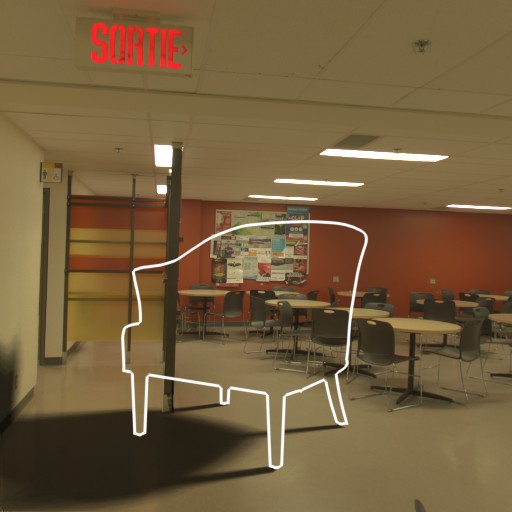} & \includegraphics[width=\mywidth]{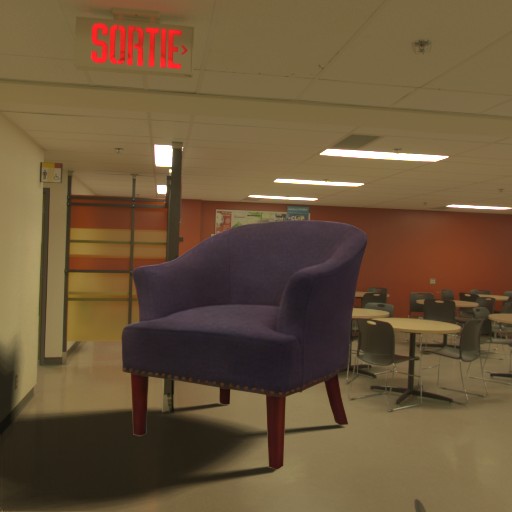} & \includegraphics[width=\mywidth]{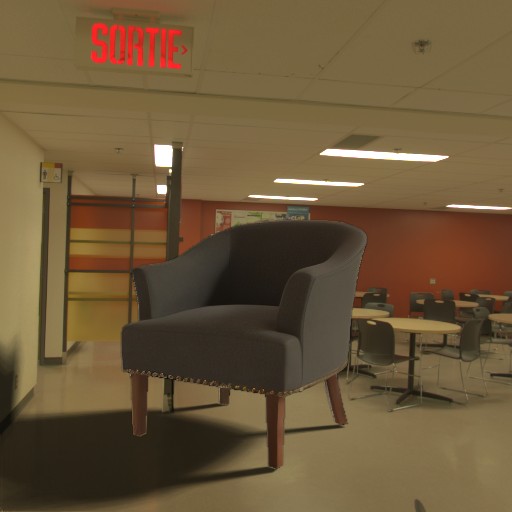} & \includegraphics[width=\mywidth]{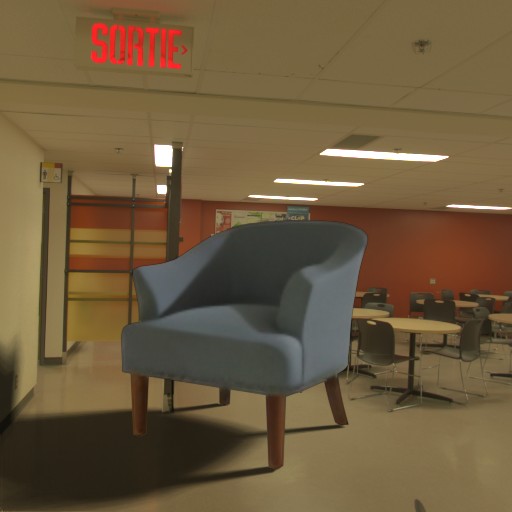} & \includegraphics[width=\mywidth]{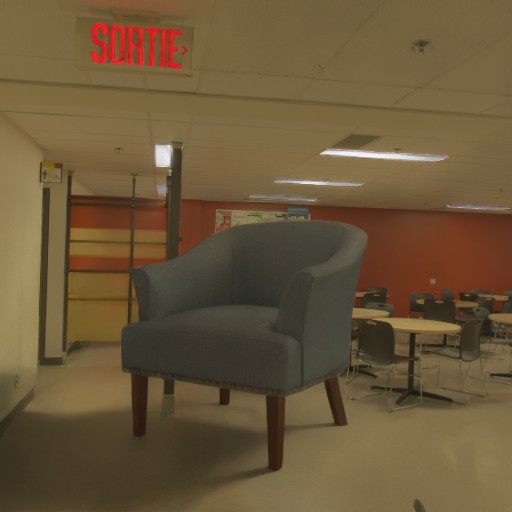} & \includegraphics[width=\mywidth]{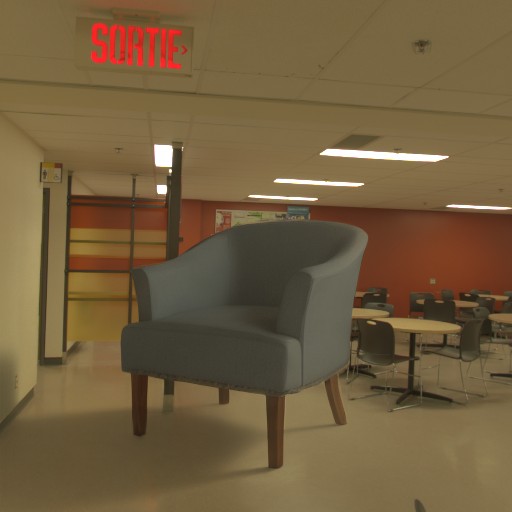} & \includegraphics[width=\mywidth]{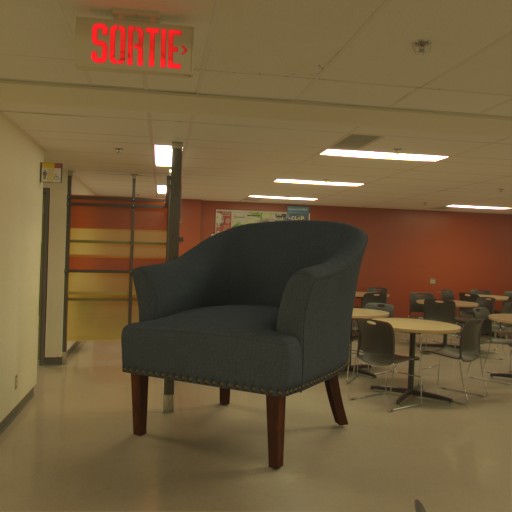}\\  
\includegraphics[width=\mywidth]{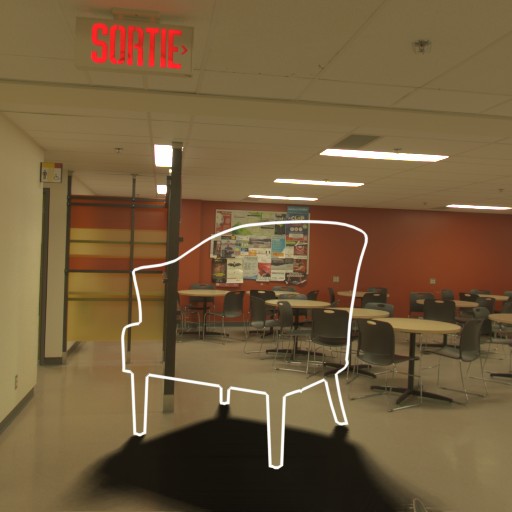} & \includegraphics[width=\mywidth]{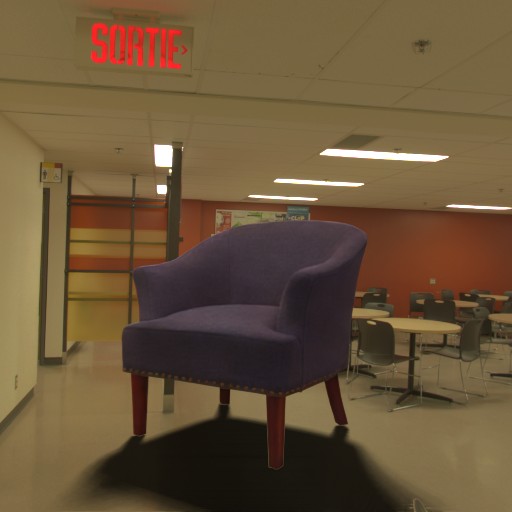} & \includegraphics[width=\mywidth]{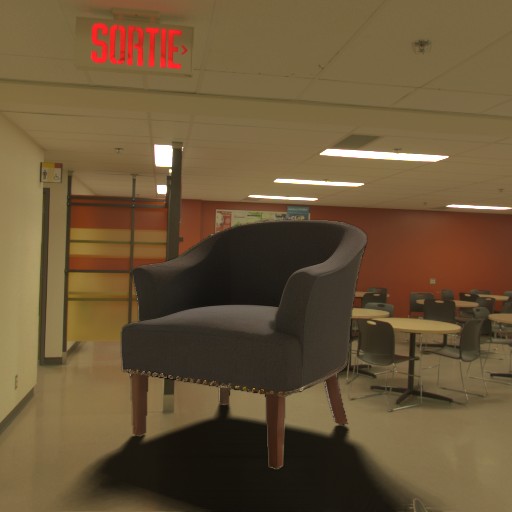} & \includegraphics[width=\mywidth]{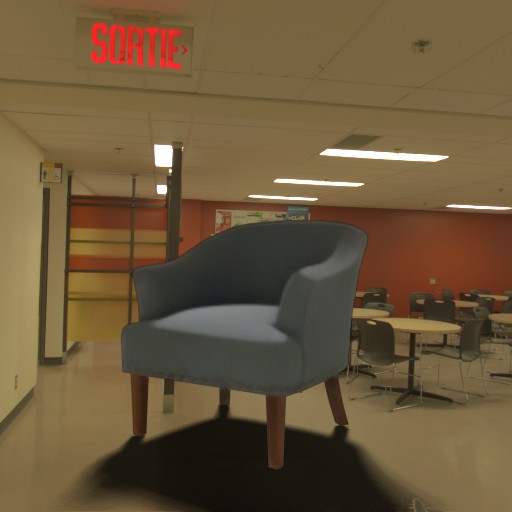} & \includegraphics[width=\mywidth]{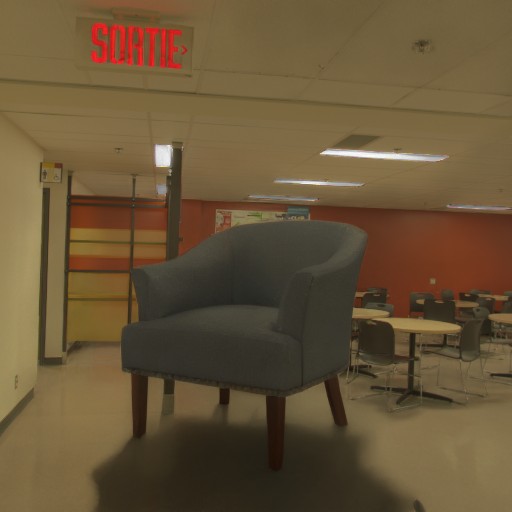} & \includegraphics[width=\mywidth]{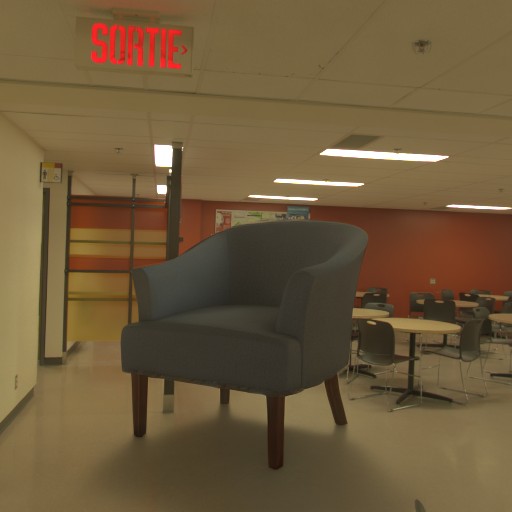} & \includegraphics[width=\mywidth]{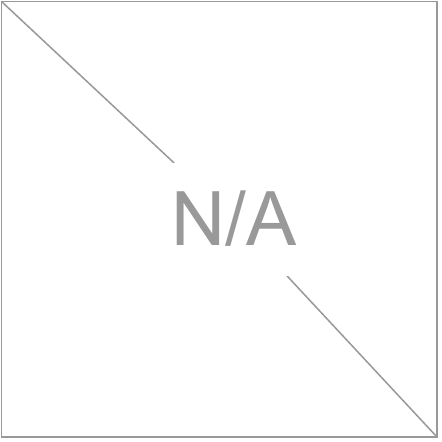}\\  
\includegraphics[width=\mywidth]{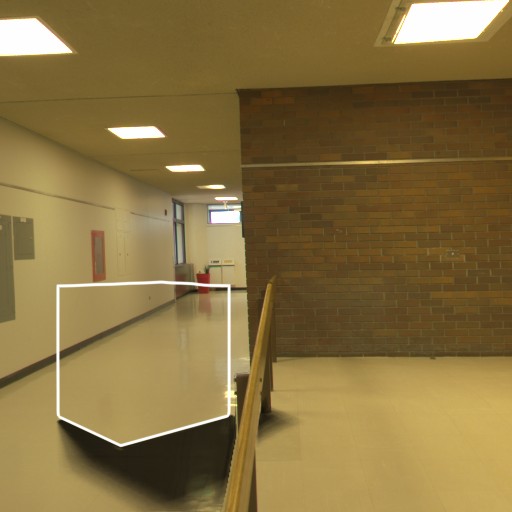} & \includegraphics[width=\mywidth]{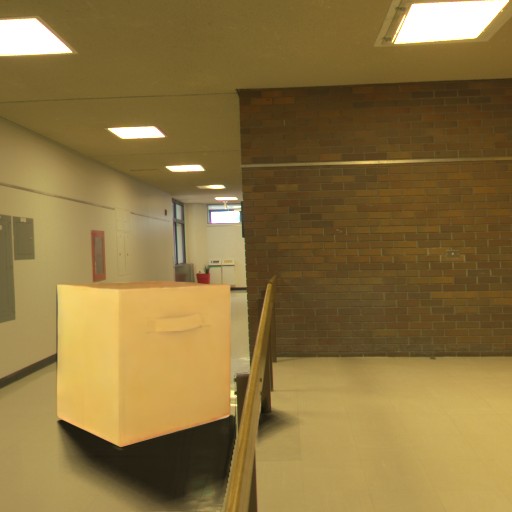} & \includegraphics[width=\mywidth]{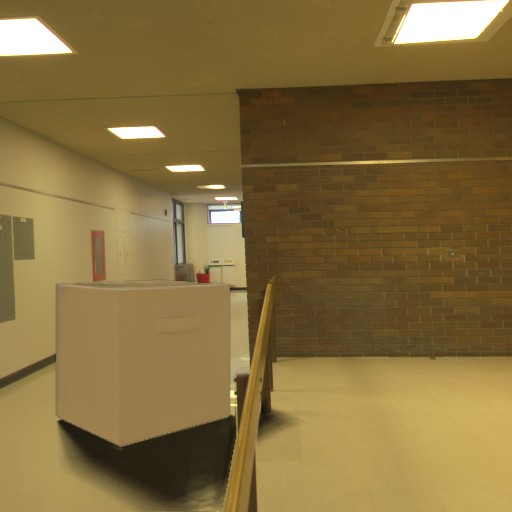} & \includegraphics[width=\mywidth]{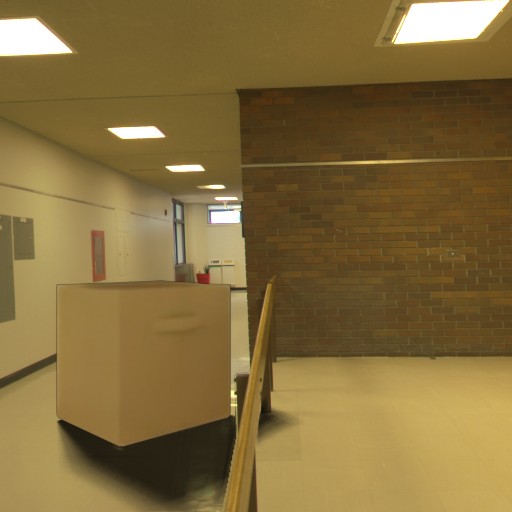} & \includegraphics[width=\mywidth]{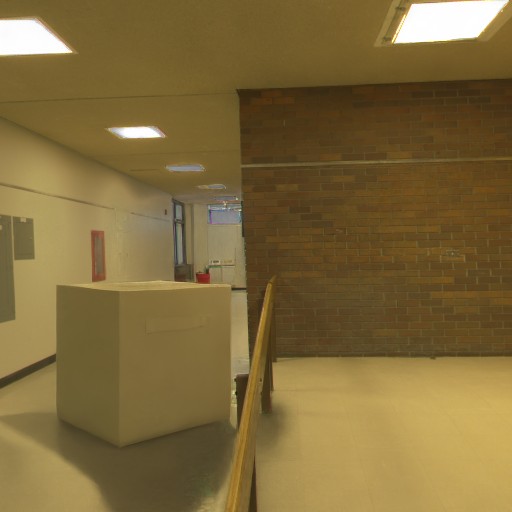} & \includegraphics[width=\mywidth]{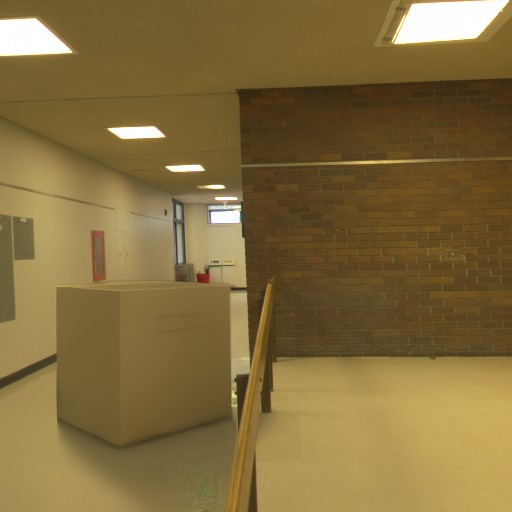} & \includegraphics[width=\mywidth]{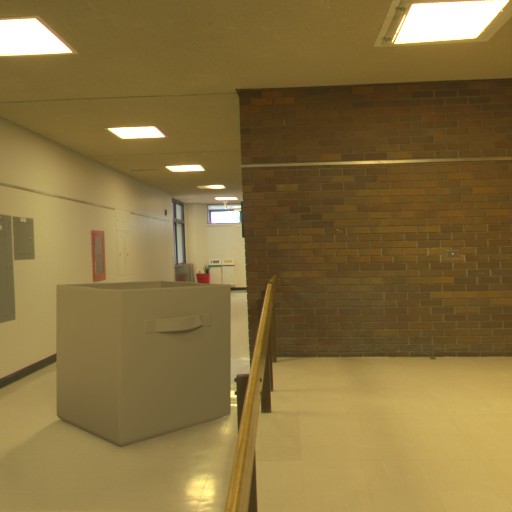}\\  
\includegraphics[width=\mywidth]{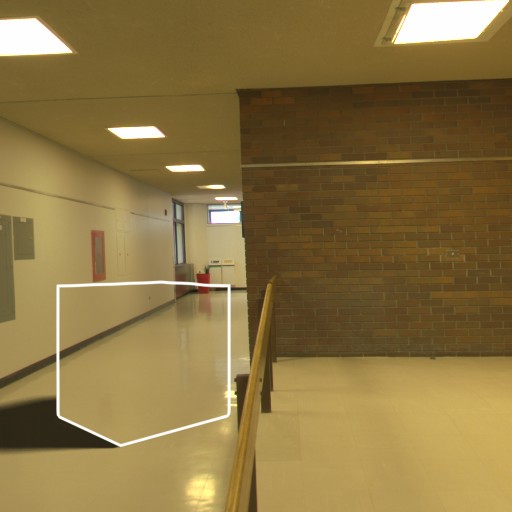} & \includegraphics[width=\mywidth]{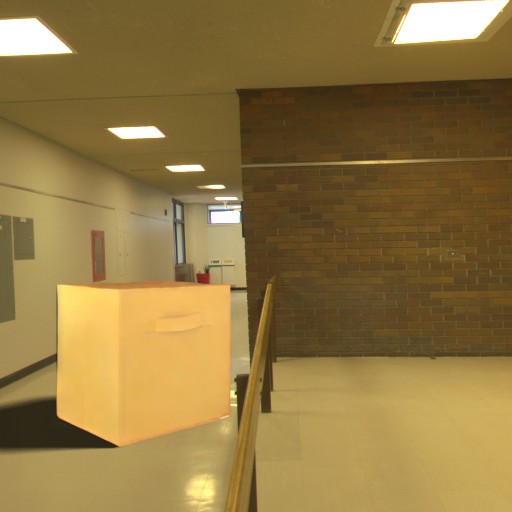} & \includegraphics[width=\mywidth]{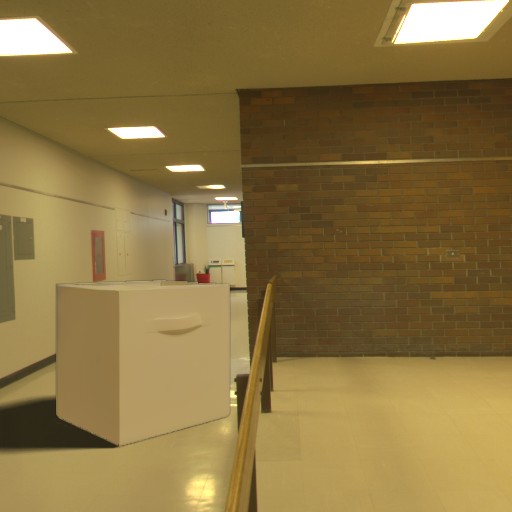} & \includegraphics[width=\mywidth]{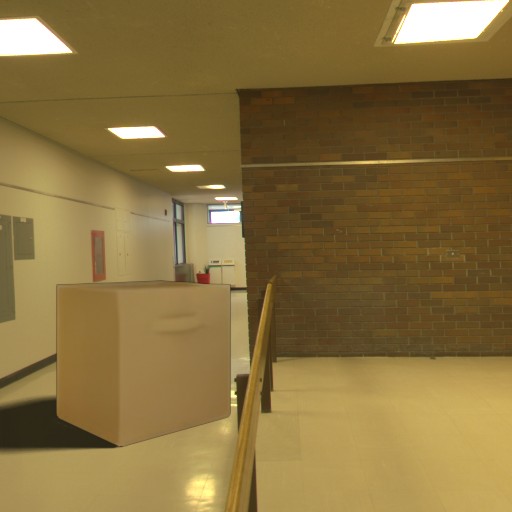} & \includegraphics[width=\mywidth]{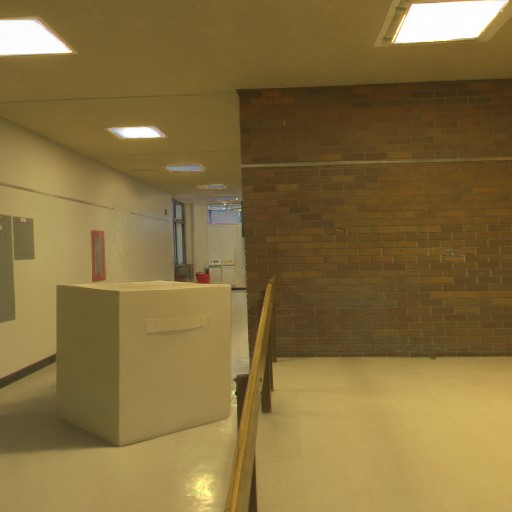} & \includegraphics[width=\mywidth]{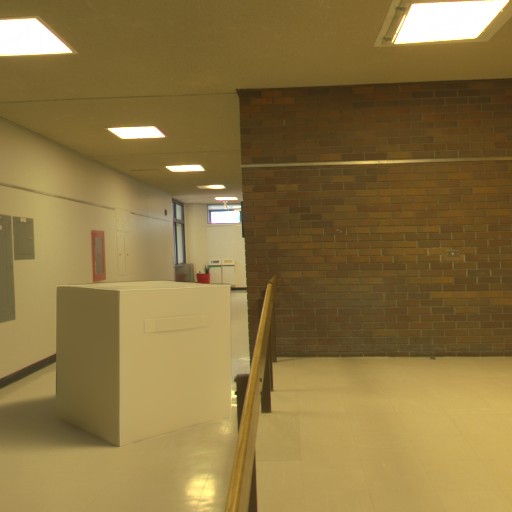} & \includegraphics[width=\mywidth]{figs/NA.pdf}\\  
\includegraphics[width=\mywidth]{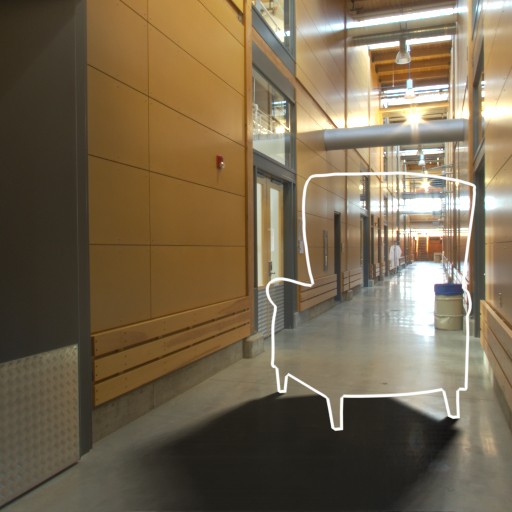} & \includegraphics[width=\mywidth]{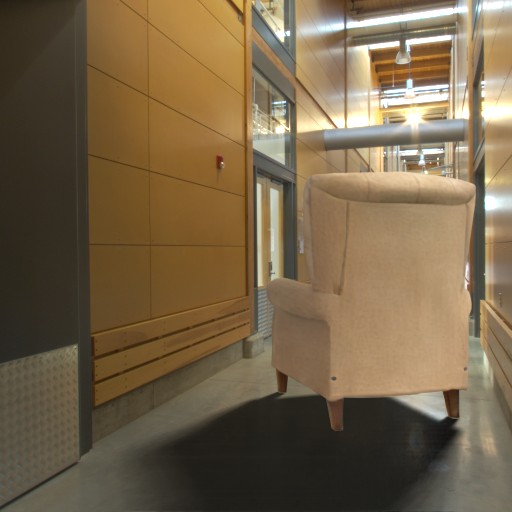} & \includegraphics[width=\mywidth]{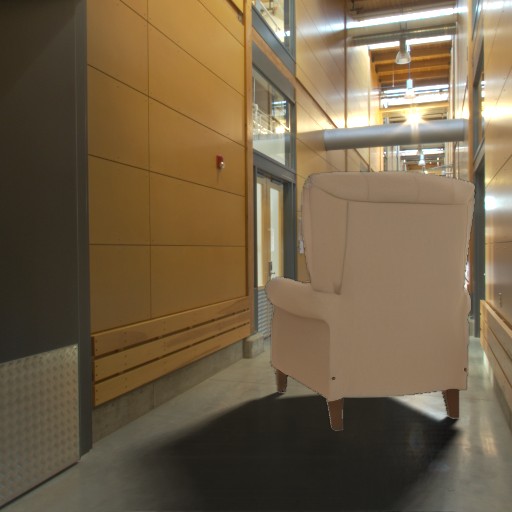} & \includegraphics[width=\mywidth]{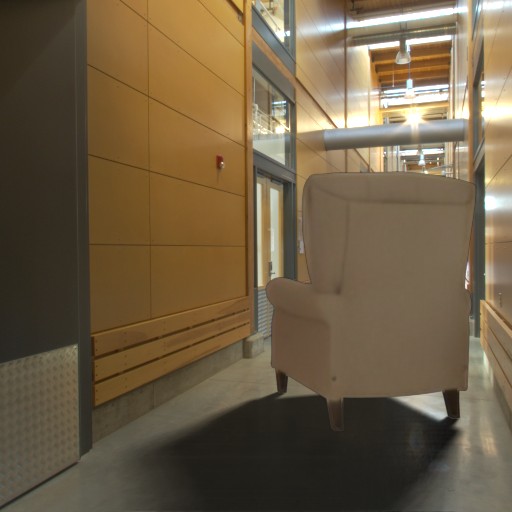} & \includegraphics[width=\mywidth]{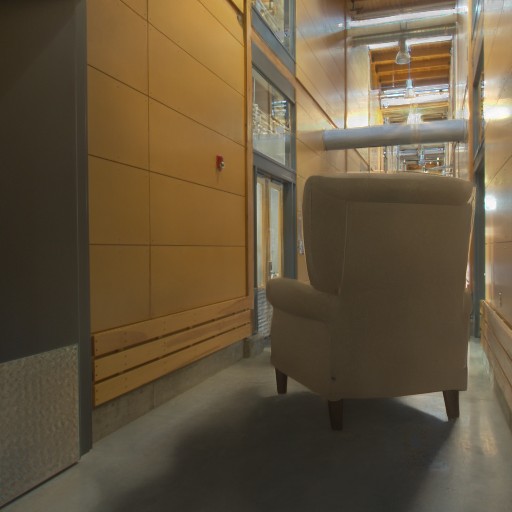} & \includegraphics[width=\mywidth]{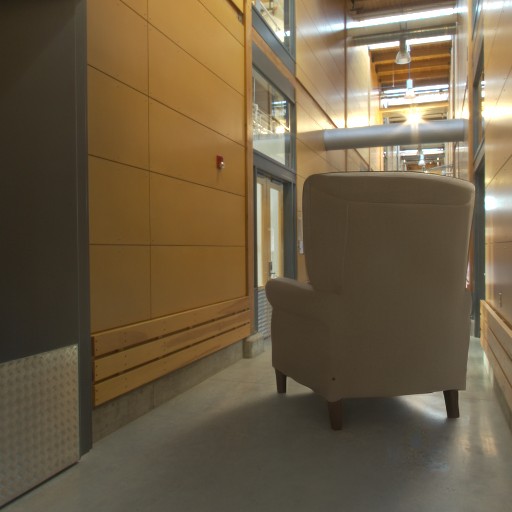} & \includegraphics[width=\mywidth]{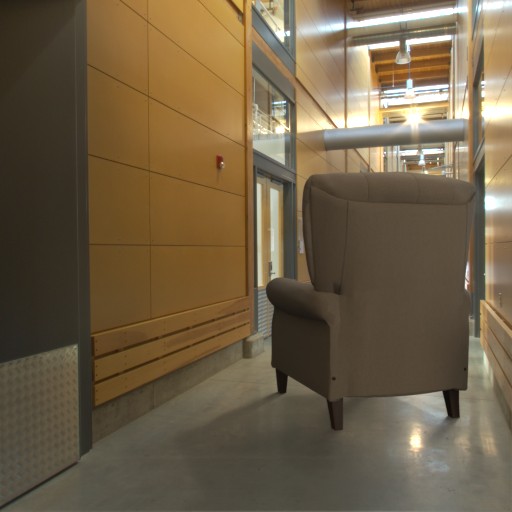}\\  
\includegraphics[width=\mywidth]{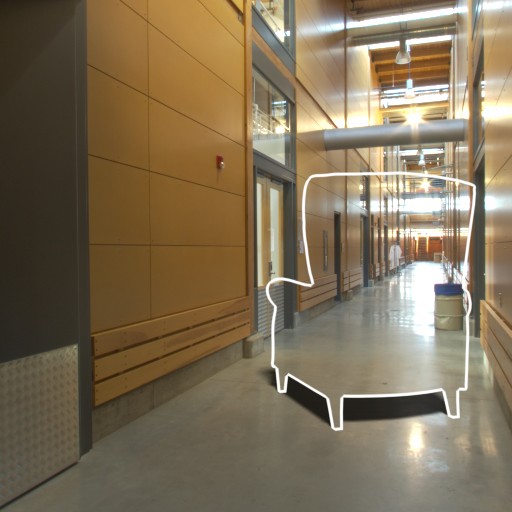} & \includegraphics[width=\mywidth]{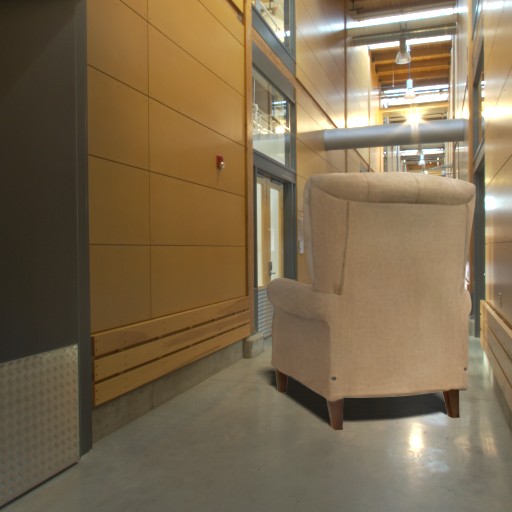} & \includegraphics[width=\mywidth]{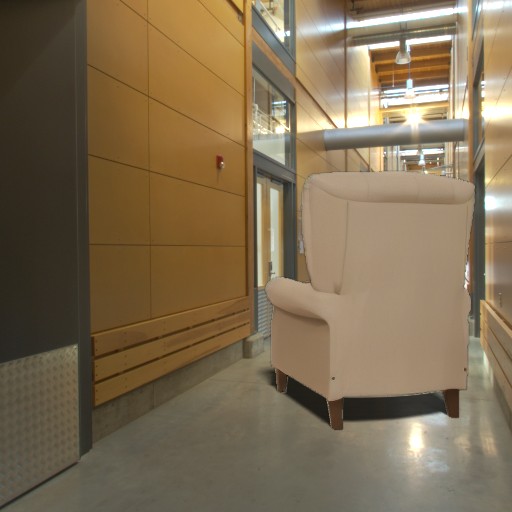} & \includegraphics[width=\mywidth]{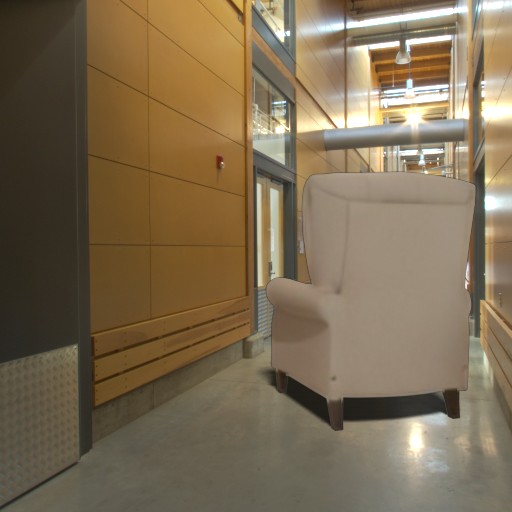} & \includegraphics[width=\mywidth]{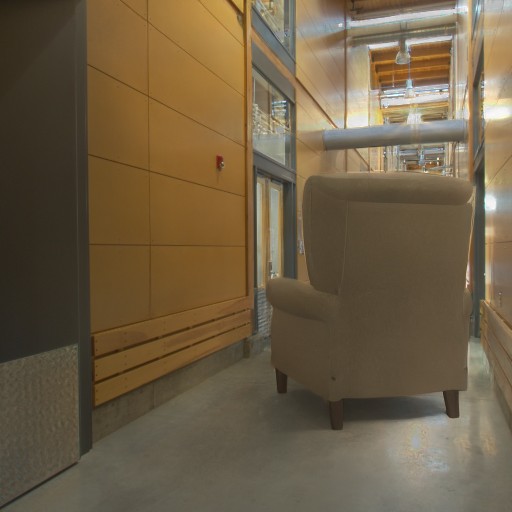} & \includegraphics[width=\mywidth]{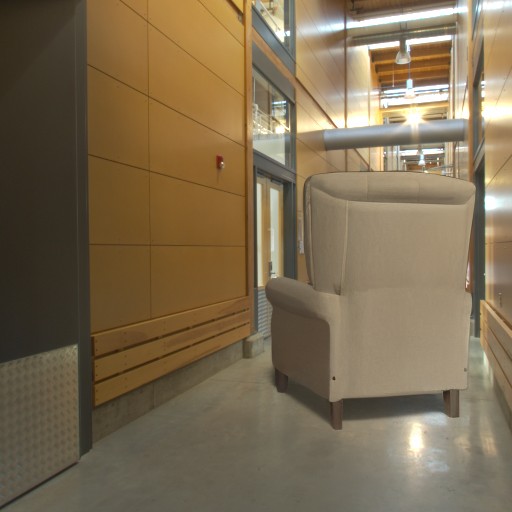} & \includegraphics[width=\mywidth]{figs/NA.pdf}\\  
\includegraphics[width=\mywidth]{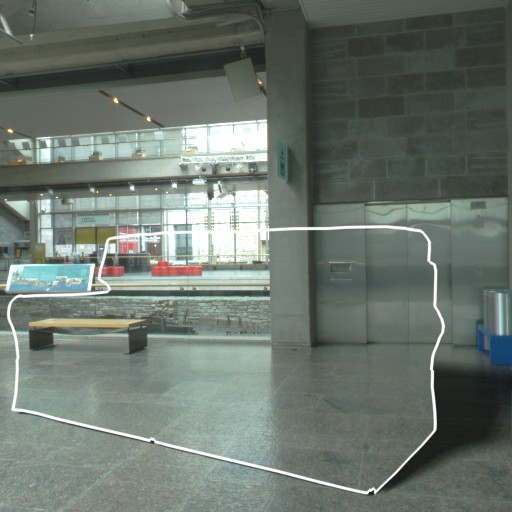} & \includegraphics[width=\mywidth]{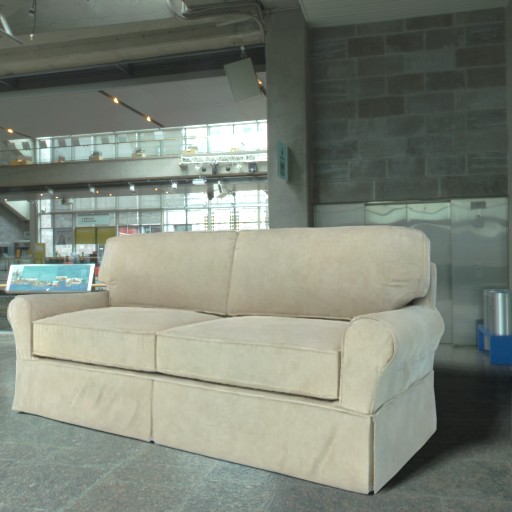} & \includegraphics[width=\mywidth]{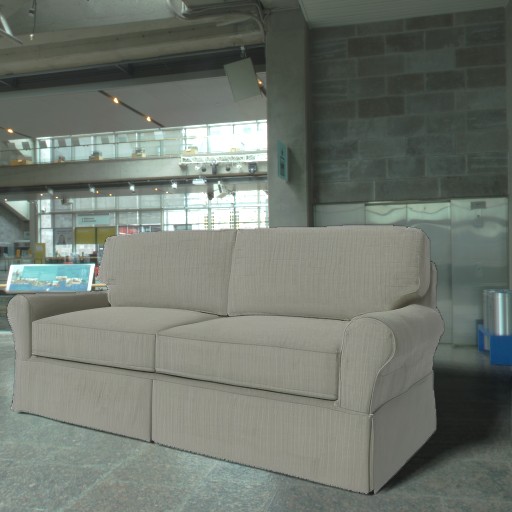} & \includegraphics[width=\mywidth]{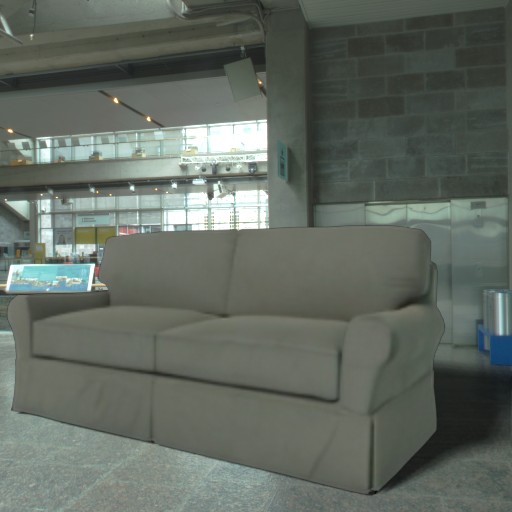} & \includegraphics[width=\mywidth]{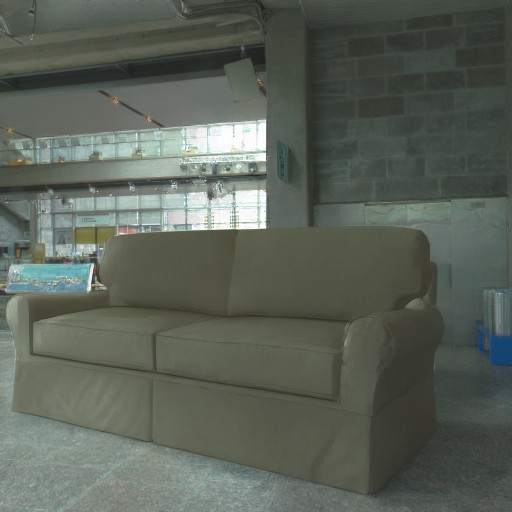} & \includegraphics[width=\mywidth]{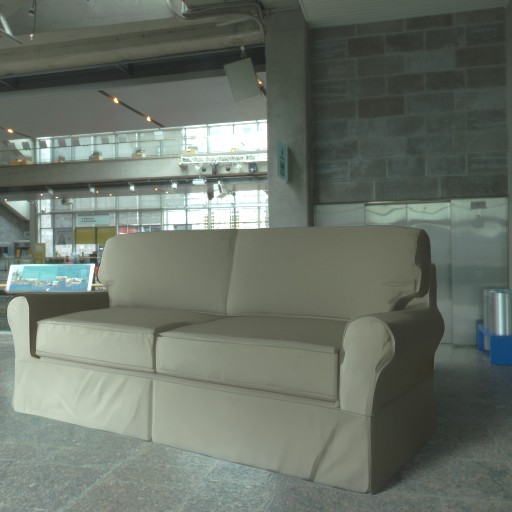} & \includegraphics[width=\mywidth]{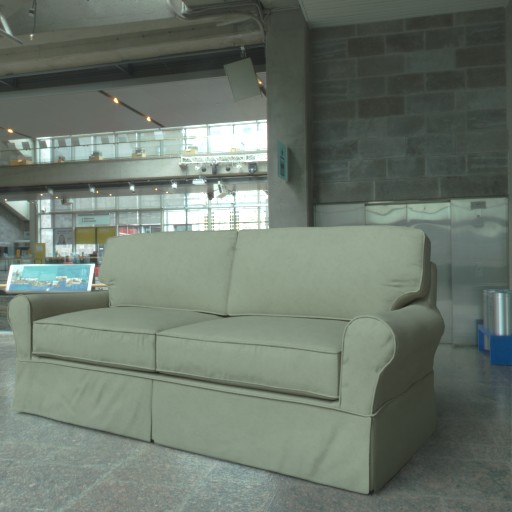}\\  
\includegraphics[width=\mywidth]{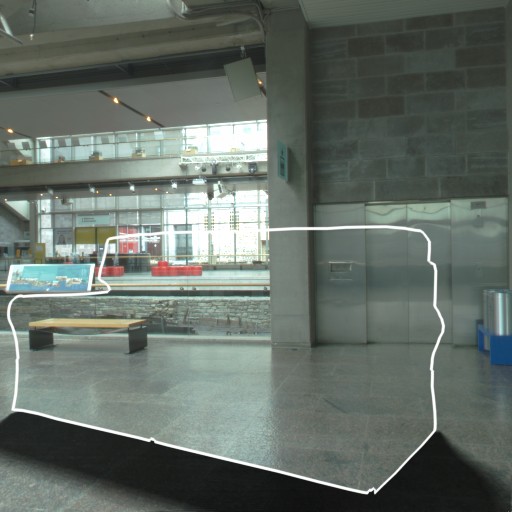} & \includegraphics[width=\mywidth]{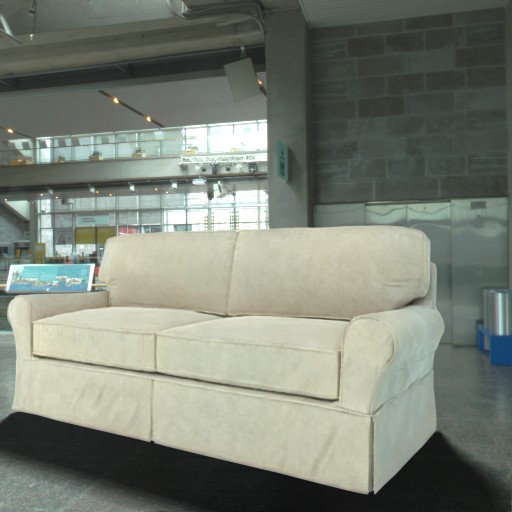} & \includegraphics[width=\mywidth]{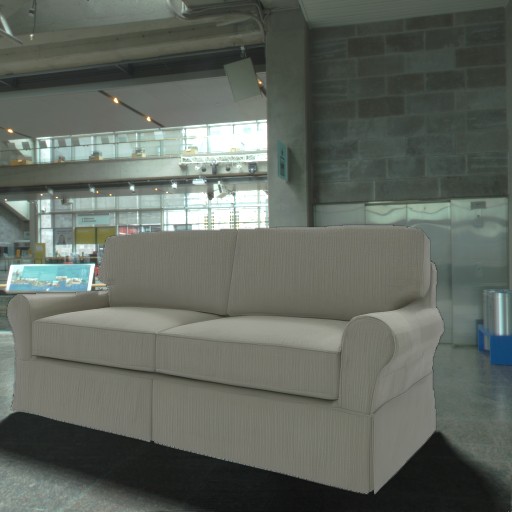} & \includegraphics[width=\mywidth]{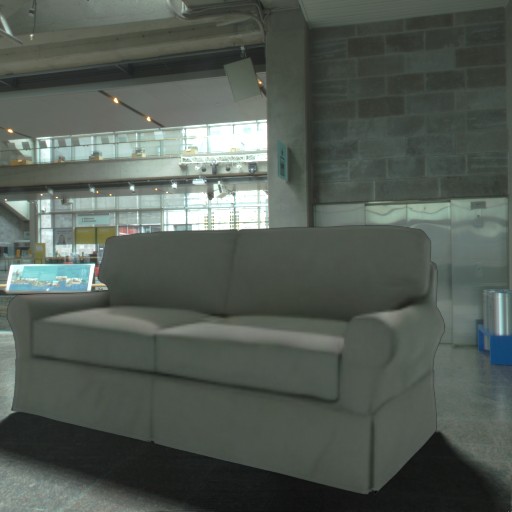} & \includegraphics[width=\mywidth]{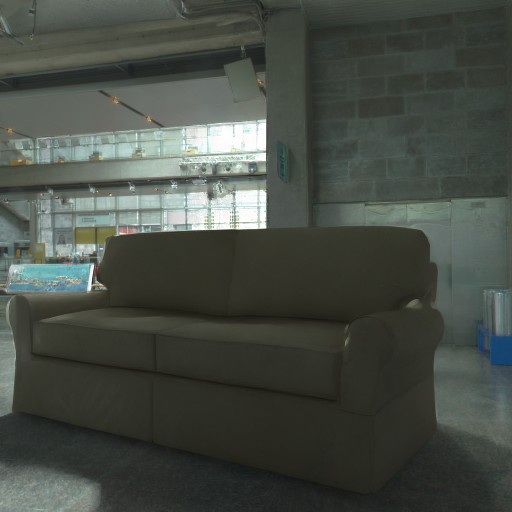} & \includegraphics[width=\mywidth]{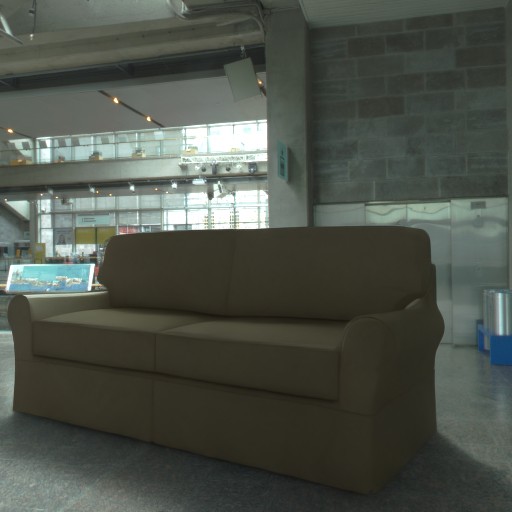} & \includegraphics[width=\mywidth]{figs/NA.pdf}\\  
Guiding shadow & IC-Light~\cite{zhang2025iclight} & DiLightNet~\cite{zeng2024dilightnet} & Neural Gaffer~\cite{jin2024neural_gaffer} & ZeroComp+SDEdit & Ours & Ground truth
 \end{tabular}

%% file: sec/evaluation.tex
\section{Evaluation on 3D object compositing}
\label{sec:evaluation}

Evaluating composition methods where lighting conditions can be controlled is challenging, as the quality of the results must be assessed across multiple lighting condition for the same scene.
To address this, we build over the evaluation setup of ~\cite{zhang2025zerocomp} where 3D objects are realistically rendered and composited into background images using a physics-based renderer and ground-truth lighting extracted from HDR panoramas. We evaluate two scenarios: (1)~``reference-based'', which preserves the original scene lighting to assess the ability of methods to produce images close to the ground truth, and (2) ``user-controlled'', where the original lighting is modified (e.g., dominant light rotation) to simulate user-specified lighting conditions, requiring methods to adapt while maintaining a perceptually pleasing render. In this section, we describe how these scenarios are used for quantitative evaluation and a user study to assess the realism and controllability of the methods.

%In this section, we quantitatively evaluate \method against other methods that support lighting control for 3D object composition in RGB images. Additionally, we conduct user studies to assess human perception and compare its performance with competing approaches.

\subsection{Evaluation dataset}
\label{sec:eval-dataset}

%We start from the evaluation dataset proposed in ZeroComp~\cite{zhang2025zerocomp}, where high-quality 3D models from the Amazon Berkeley Objects dataset~\cite{collins2022abo} are inserted into 50° field of view background images extracted from panoramas in the Laval Indoor HDR dataset~\cite{gardner2017sigasia}. The authors of \cite{zhang2025zerocomp} graciously provided us with reproducible Blender scene information, which we extend in three ways. 
We create datasets for the two scenarios using the Blender scenes graciously provided by the authors of~\cite{zhang2025zerocomp}.
We first improve the 3D object rendered shadow by replacing the planar shadow catcher by a scene mesh using estimated depth from Depth Anything V2~\cite{depth_anything_v2}. We also improve the ground truth intrinsic maps with antialiased versions. 

%First, we employ a mesh-based (instead of planar) shadow catcher, obtained by back-projecting the background depth estimated using Depth Anything V2~\cite{depth_anything_v2}. Second, we annotate a simple parametric dominant light source and render its associated shadow using shadow mapping. Third, object intrinsics are rendered with anti-aliased alpha mattes, allowing for more realistic compositing results. We produce two versions of the dataset, each intended for evaluating different aspects of our method.

\myparagraph{``Reference-based'' scenario.} To assess the capacity of the methods to generate a composite in the ground truth lighting conditions, this scenario uses renders with the depth-warped HDR as in~\cite{zhang2025zerocomp} to ensure lighting fidelity around the inserted object. We extract the dominant light direction and use it to render the guiding shadow using blender's EEVEE rasterization-based rendering engine, which employs real-time shadow mapping~\cite{eisemann2011realtime}.
A total of 210 images featuring various objects are rendered, with example samples shown in \cref{fig:qual-usercontrol}, in the ``ground truth'' column. This dataset is intended for quantitative evaluation (\cref{sec:eval-results}).

%This version uses the dominant light source direction in the scene panorama. A total of 210 images featuring various objects are rendered, with example samples shown in \cref{fig:qual-usercontrol}, in the ``ground truth'' column. We filter out samples where the background depth estimation failed. This dataset is intended for quantitative evaluation (\cref{sec:eval-results}), allowing the ground truth of the inserted object to be used to compute metrics. For the ground truth renders, we use a depth-warping strategy on the HDR panorama as in~\cite{zhang2025zerocomp} to account for spatially varying lighting around the object position. 
%ground-truth HDR map can be used to compute relevant metrics.
% filtered out 3 examples from 213
\myparagraph{``User-controlled'' scenario.} This dataset simulates a scenario where a user specifies a desired light position. 8 lighting directions are defined, spaced at $45^\circ$ increments in azimuth around the object, and used to create the guiding shadows, again using blender's EEVEE engine. The same set of background and object combinations as in the ``reference-based'' version is used, resulting in a total of 1,680 images. This dataset is used in the user study (\cref{sec:eval-user-study}) to evaluate the perceptual accuracy of renders under modified lighting conditions.

Note that these evaluation datasets are independent of the datasets used to train both the backbones employed in \method (\cref{sec:eval-specifics}) and the baselines (\cref{sec:baselines}).

% \todo{1704} light control position, plus 213 light direction matching the ground truth light direction. 

% \todo{shouldn't we emphasize all the hard work that went into generating controllable light directions, with light direction estimates, etc.? That could be part of the dataset contribution}

% \todo{Distinguish between 2 datasets: 1) reference-based (dominant light direction, so we have a simulated GT)---allows us to compute quantitative metrics; 2) control-based (light directions are specified manually), here we have no GT }

% 1) adding explicit light control, where a distant parametric light is moved around the object; 2

% \input{figs/figure_qualitative_results}

\subsection{\method specifics}
\label{sec:eval-specifics}

As mentioned in \cref{sec:background-diffusion-renderers}, our proposed \method readily applies to existing diffusion-based neural renderers, such as ZeroComp~\cite{zhang2025zerocomp} or RGB$\leftrightarrow$X~\cite{zeng2024rgb}. Here, we provide details on ZeroComp, see supp. for RGB$\leftrightarrow$X.

% \myparagraph{ZeroComp~\cite{zhang2025zerocomp}} provides a maskable shading map as input to a ControlNet-based neural renderer. Here, we use their approach directly and mask out the shading on both the object and the shadow region. 
% \myparagraph{RGB$\leftrightarrow$X~\cite{zeng2024rgb}} provides a 

% \myparagraph{ZeroComp backbone}
% We train ZeroComp following the procedure outlined in~\cite{zhang2025zerocomp}, and use the same conditioned intrinsic estimators at inference time:  
We use the variant trained on Openrooms~\cite{li2020openrooms} provided by the authors. The masked shading map is obtained by dividing the background image by the albedo, and by masking out the shading on both the object and shadow regions. Given its superior results, we adopt this method as our default model for 3D object compositing. To extract intrinsics from real images, we use the same conditioned intrinsic estimators as in \cite{zhang2025zerocomp}, namely StableNormal~\cite{ye2024stablenormal} for normals, IID~\cite{kocsis2023iid} for albedo, and ZoeDepth~\cite{bhat2023zoedepth} for depth.

%\myparagraph{Light control} Since \method is conditioned by the shadows of the object, we use a distant spherical light with a radius and an intensity chosen to approximate the properties of the spherical gaussian light source. The shadow is rendered with EEVEE as mentioned in \cref{sec:eval-specifics}.

%Using a real-time shadow mapping method, instead of ray-tracing, allows for real-time shadow editing. 
 
% \myparagraph{Estimating a background mesh.}
% Recently, there has been a big increase in the quality of monocular depth estimation methods, with several new methods with increasing reconstruction quality. We find that the relative version of Depth Anything v2 yields particularly good results when we reconstruct a mesh. We apply an affine transform to the object's depth to match the background 
% \todo{zerocomp also did that, but didn't do the affine transform in disparity space, only in depth space}. In our case, we find an affine transform in inverse-depth to apply to the background, in order to obtain a valid 3D mesh. \todo{equation LSQ}

\begin{table}[t]
    \centering
    \setlength{\tabcolsep}{1pt}
    \small
    \begin{tabular}{lccccc}
    \toprule
    Method & PSNR$_\uparrow$  & SSIM$_\uparrow$  & RMSE$_\downarrow$ & MAE$_\downarrow$ & LPIPS$_\downarrow$ \\
    \midrule
    \multicolumn{6}{l}{\emph{Light-conditioned methods}} \\
    DiLightNet~\cite{zeng2024dilightnet} & 24.67 & 0.948 & 0.064 & 0.022 & 0.042 \\
    Neural Gaffer~\cite{jin2024neural_gaffer} & \cellcolor{orange!25}28.44 & \cellcolor{orange!25}0.963 & \cellcolor{orange!25}0.042 & \cellcolor{orange!25}0.015 & \cellcolor{orange!25}0.038 \\
    \midrule
    \multicolumn{6}{l}{\emph{Shadow-conditioned methods}} \\ 
    IC-Light~\cite{zhang2025iclight} & 26.87 & 0.959 & 0.054 & 0.019 & 0.040 \\
    ZeroComp+SDEdit~\cite{meng2022sdedit} & 26.00 & 0.938 & 0.053 & 0.025 & 0.079 \\
    % \method (RGB$\leftrightarrow$X) & 26.56 & 0.955 & 0.051 & 0.018 & 0.042 \\
    \method & \cellcolor{red!25}30.68 & \cellcolor{red!25}0.974 & \cellcolor{red!25}0.033 & \cellcolor{red!25}0.012 & \cellcolor{red!25}0.030 \\
    \bottomrule
    \end{tabular}
    \caption{Quantitative results obtained by conditioning the methods on the ground truth dominant light direction. All metrics are computed on our ``reference-based'' evaluation dataset against the ground truth on the full image (see supp. for foreground- and background-only metrics). \method surpasses all baselines. Results are color coded by \colorbox{red!25}{best} and \colorbox{orange!25}{second-}best.}
    \label{tab:quant_results}
\end{table}

\subsection{Baselines}
\label{sec:baselines}

We compare \method with two types of baselines, that is methods that are conditioned on: the guiding shadow (like \method); or an explicit lighting representation, through a $360^\circ$ environment map. We now describe per-method details. OpenImageDenoise~\cite{OpenImageDenoise} is also used to denoise the results of all methods to limit the impact of rendering noise present in the synthetic training data of some models.

%We compare \method against recent diffusion-based methods that support light control: DiLightNet~\cite{zeng2024dilightnet}, IC-Light~\cite{zhang2025iclight}, and Neural Gaffer~\cite{jin2024neural_gaffer}, using their public implementations. While these baselines can relight objects, none are capable of generating a shadow around the object. For fairness, the physically-based renderer Cycles is therefore used to cast a realistic shadow. OpenImageDenoise~\cite{OpenImageDenoise} is also used to denoise the results of all methods. We begin by describing how light control is given to baselines, then proceed to describe per-method specifics.

% \begin{reviseEnv}
\subsubsection{Shadow-conditioned methods}
These methods, like ours, use the user-specified guiding shadow $\mathbf{m}_\text{shw}$ as input (obtained with EEVEE, see \cref{sec:eval-dataset}). 
% They do not require access to a lighting environment map.

\myparagraph{ZeroComp~\cite{zhang2025zerocomp}+SDEdit~\cite{meng2022sdedit}.} 
We create a strong baseline by first running ZeroComp with the shading of the object masked out, and composite the guiding shadow $\mathbf{m}_\text{shw}$ over the background shading intrinsic map. To blend the shadows, we perform a refinement step, inspired by SDEdit~\cite{meng2022sdedit}, by noising this prediction for 50\% of the timesteps, and re-running ZeroComp for the remaining timesteps, this time with the shading of the whole image masked out.

\myparagraph{IC-Light~\cite{zhang2025iclight}.} 
% Lous-Étienne's documented implementation:
% https://github.com/LVSN-Compositing/IC-Light_eval
We evaluate the background-conditioned model of IC-Light, where the background image with the shadow is used for conditioning, with the object lit by a constant ambient lighting as input. No prompts are used.

\subsubsection{Light-conditioned methods}
Object relighting methods accept an explicit lighting representation as input in the form of a $360^\circ$ environment map, which we generate using the parametric light direction (see supp.). Unfortunately, these methods focus solely on object relighting---they do not create any cast shadows outside the object. To prevent this absence of shadows from severely hampering the realism of their results, we composite the guiding shadow onto the background image. 

% Since using the raw output would severly hamper the consistency of the output with the target background, we instead composite the background

% For the latter, existing diffusion-based relighting methods conditionned on lighting are unable to produce shadows on the original background. Since using the raw output would severly hamper the consistency of the output with the target background, we instead composite the background with a shadow cast by the physically-based renderer Cycles (see supp. for the raw method outputs). 

% Since these methods either inpaint a background or don't generate one, we replace the background with the shadow cast by the physically-based renderer Cycles. For each scene and lighting direction, a simple environment map containing the dominant light source is fed to the methods, see supplementary for details.

\myparagraph{DiLightNet~\cite{zeng2024dilightnet}} 
% Lous-Étienne's documented implementation:
% https://github.com/LVSN-Compositing/DiLightNet_eval
requires a prompt describing the object, which we obtain by feeding the object lit by constant ambient lighting and composited on a black background to BLIP-2~\cite{li2023blip2} with the prompt ``Putting aside the black background, this object is a''. 
% We also feed into the method the ground truth foreground and the spherical gaussian and ambient lighting estimated from the average background color as an HDR environment map. 
% To generate the results, depth and radiance hints are required. 
DiLightNet also requires depth and radiance hints: we obtain depth from the 3D model of the object and render the radiance hints using the input environment map as outlined in~\cite{zeng2024dilightnet}. The rendered object is then composited with the background. 
% \todo{say that the method also requires the image of obj lit by constant ambient}

\myparagraph{Neural Gaffer~\cite{jin2024neural_gaffer}.}
% Louis-Étienne's documented implementation:
% https://github.com/LVSN-Compositing/Neural_Gaffer_eval
The object lit by constant ambient lighting is composited on a white background before feeding it to the network, as expected by the method. Since Neural Gaffer is trained at a lower resolution ($256\times256$), its result is upsampled to $512\times512$ then composited on the full resolution background. We noticed Neural Gaffer tends to generate noticeable white boundary artifacts---we remove them through an erosion operation to avoid overly penalizing it.  %To mitigate the white boudary artifacts appear around the object, we replace the outer 5 pixels of the object edge with the 6th pixel inward \todo{clarify}.

\subsection{Experimental results}
\label{sec:eval-results}

\Cref{tab:quant_results} presents quantitative results comparing the various baselines (c.f. \cref{sec:baselines}) with \method (c.f. \cref{sec:eval-specifics}) on our ``reference-based'' dataset, where the ground truth is available (\cref{sec:eval-dataset}). Here, all methods are conditioned on the ground truth dominant light direction using their specific light parametrization (\cref{sec:baselines}). \Cref{fig:qual-usercontrol} shows qualitative results for each method. We observe that \method with the ZeroComp backbone outperforms the baselines on quantitative metrics and produces visually superior results.

\begin{figure}[t]
    \centering
    \footnotesize
    \includegraphics[width=1.0\linewidth]{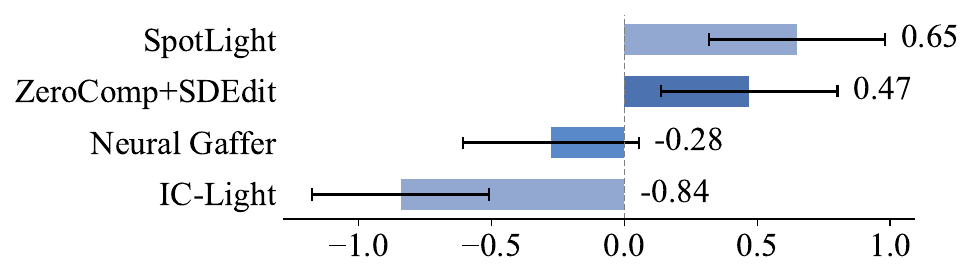} \\
    (a) Overall realism study \\
    \includegraphics[width=1.0\linewidth]{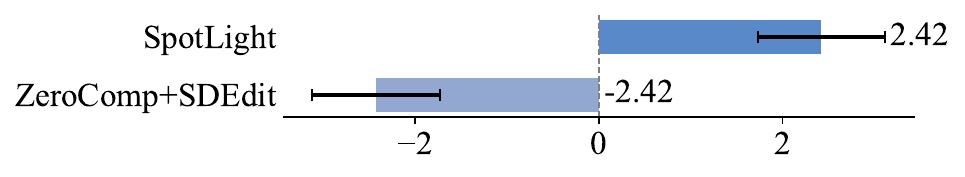} \\
    (b) Lighting control study
    \vspace{-5pt}
    \caption{User study results, displaying the Thurstone Case V $z$-scores along with the 95\% confidence intervals. \method was preferred over baselines by users both in terms of (a) overall realism and (b) lighting control.}
    % TODO: use set2 colors
    \label{fig:user-study}
    \vspace{-5pt}
\end{figure}

\subsection{User studies}
\label{sec:eval-user-study}

Two user studies were conducted to assess the perceptual realism and controllability of \method compared to baselines. In both cases, the Thurstone Case V Law of Comparative Judgement~\cite{thurstone2017} is used to obtain a $z$-score for each method, where a higher value indicates higher human preference. Here, we give an overview of those user studies. Refer to the supp. for additional details and studies.

\paragraph{User study I: Overall realism}
Here, we use the ``user-controlled'' version of the dataset (see \cref{sec:eval-dataset}) to evaluate the realism of results obtained by conditioning on a specific lighting direction, even if it is not aligned with the scene lighting conditions. We compare \method against the three baselines with the best PSNR from \cref{tab:quant_results}, namely Neural Gaffer~\cite{jin2024neural_gaffer}, IC-Light~\cite{zhang2025iclight}, and ZeroComp~\cite{zhang2025zerocomp}+SDEdit~\cite{meng2022sdedit}.
Since no ground truth is available, we design a two-alternative forced choice user study, where results obtained by two methods are shown side by side. The participants are asked to ``Click on the image that looks the most realistic, by considering both the appearance of the object and its shadows.'' For each pair of methods, a set of 20 random images are chosen (we ensure images where shadows are clearly visible are selected), resulting in 120 pairs of images observed by each participant. We randomly shuffle the 120 pairs for each user, and the left-right order. Three sentinel images were added to discard observers that did not understand the task. Examples are shown in \cref{fig:qual-usercontrol}.

User study I was completed by $N=35$ observers. \Cref{fig:user-study}a shows that our method is the favorite, indicating that \method achieves the highest realism, with a statistically significant difference to Neural Gaffer and IC-Light. 

% \revise{We further ran additional user studies that evaluate separately the shadow realism, the shading realism and the level of control, which can be found in the supplementary materials.} \todo{talk about the outcome of those studies?}

\myparagraph{User study II: lighting control} Since the goal of our method is to give full artistic control over the object relighting, we design a second two-alternative forced choice user study to evaluate lighting controllability separately from the realism. The users were shown two videos side by side and were asked to ``Select the video where the light direction has the most impact on the lighting of the object, by considering both the appearance of the object and its shadows.''. The videos show a virtual object relit by a rotating light source (see supp.).
Here, we compared the two methods with the highest realism according to user study I: \method and ZeroComp+SDEdit. Since this task aims to evaluate the local relighting within the object mask only, we replace the shadowed background with the output from our method.

User study II was completed by $N=8$ observers. As seen in \cref{fig:user-study}b, \method achieves a statistically significant improvement in performance over ZeroComp+SDEdit, despite the low number of participants, which we attribute to the use of our local guidance strategy. Please see the supp. for more user study results and experimental details.

% Since our method's error bars don't overlap with the baselines, we confirm that our results are statistically significant \todo{p-value?}.

% \begin{figure}[t]
%     \centering
%     \includegraphics[width=1.0\linewidth]{figs/user_studies/thurstone_ICCV User Study (control, shading only).pdf}
%     \caption{Lighting control user study. The Thurstone Case V $z$-scores are displayed, along with the 95\% confidence intervals overlayed on each bar. Our method was preferred over the most realistic baseline, baselines and achieves statistical significance against both light-conditioned methods. \todo{sub-figures}}
%     % TODO: use set2 colors
%     \label{fig:user-study-plot-control}
% \end{figure}

\subsection{Effect of parameter selection}
\label{sec:ablations}

% \todo{Discuss \cref{fig:effect-light-size-position} first (once it's ready)}

Our method uses two tunable parameters: the local guidance scale ($\gamma$ in \cref{eq:modified_cfg}) and the shadow blending weight ($\beta$ in \cref{eq:shadow-blending}). \Cref{tab:ablation_quantitative_results} reports metrics obtained on the ``reference-based'' dataset where the ground truth is available (c.f. \cref{sec:eval-dataset}). Although using no guidance ($\gamma=1$) results in better quantitative performance, we observe qualitatively in \cref{fig:ablation-cfg-scale} that this significantly reduces the visibility of generated shading on the inserted object, thereby reducing the impact of the desired local lighting control. Employing no shadow blending ($\beta=0$) results in similar metric values but makes shadows much less visible. Please refer to the supp. for a qualitative comparison for the $\gamma$, $\beta$ and the shadow softness (light radius) parameters.
% (refer to the supp. for a qualitative comparison).

\begin{table}[t]
    \centering
    \small
    \setlength{\tabcolsep}{1pt}
    \begin{tabular}{lccccc}
    \toprule
    Method & PSNR$_\uparrow$  & SSIM$_\uparrow$  & RMSE$_\downarrow$ & MAE$_\downarrow$ & LPIPS$_\downarrow$ \\
    \midrule
   % $\gamma=3$, $\beta=0.05$ (Ours) & 30.68 & 0.974 & 0.033 & 0.012 & 0.031 \\
    $\gamma=3$, $\beta=0.05$ (Ours) & 30.68 & 0.974 & 0.033 & 0.012 & 0.030 \\
    \midrule
    % $\gamma=1$ (no guidance) & 31.65 & 0.976 & 0.030 & 0.011 & 0.029 \\
    $\gamma=1$ (no guidance) & 31.69 & 0.976 & 0.029 & 0.011 & 0.029 \\
    % $\gamma=7$ & 28.60 & 0.966 & 0.044 & 0.015 & 0.036 \\
    $\gamma=7$ & 28.68 & 0.966 & 0.043 & 0.015 & 0.036 \\
    \midrule
    % $\beta=0$ & 30.78 & 0.974 & 0.033 & 0.012 & 0.030 \\
    $\beta=0$ & 30.81 & 0.974 & 0.032 & 0.012 & 0.029 \\
    % $\beta=0.2$ & 29.32 & 0.970 & 0.039 & 0.014 & 0.034 \\
    $\beta=0.2$ & 29.24 & 0.969 & 0.039 & 0.014 & 0.034 \\
    % \midrule
    %Neg. no shadow & 31.79 & 0.976 & 0.029 & 0.011 & 0.030 \\
    % Neg. 3D distance mask & 29.58 & 0.971 & 0.038 & 0.013 & 0.032 \\
    \bottomrule
    \end{tabular}
    \caption{Impact of parameter selection on quantitative metrics. We observe that using no guidance ($\gamma=1$), may provide better quantitative results. However, we also observe that these changes diminish the level of light control over the object. Our selected parameter combination provides good quantitative performance and adequate lighting control.}
    \label{tab:ablation_quantitative_results}
    \vspace{-5pt}
\end{table}

% \begin{figure}[!ht]
%     \centering
%     \begin{tabular}{ccc}
%         \includegraphics[width=0.3\linewidth]{example-image} &
%         \includegraphics[width=0.3\linewidth]{example-image} &
%         \includegraphics[width=0.3\linewidth]{example-image} \\
%         \includegraphics[width=0.3\linewidth]{example-image} &
%         \includegraphics[width=0.3\linewidth]{example-image} &
%         \includegraphics[width=0.3\linewidth]{example-image} \\
%         CFG ($\gamma = 3$) & RGB constrast boost & ?
%     \end{tabular}
%     \caption{Placeholder: Ablation: classifier-free guidance vs. RGB contrast amplification. Top: first lighting direction, bottom: second lighting direction.}
%     \label{fig:ablation-cfg-scale}
% \end{figure}

% \begin{figure}[!ht]
%     \centering
%     \begin{tabular}{ccc}
%         \includegraphics[width=0.3\linewidth]{example-image} &
%         \includegraphics[width=0.3\linewidth]{example-image} &
%         \includegraphics[width=0.3\linewidth]{example-image} \\
%         \includegraphics[width=0.3\linewidth]{example-image} &
%         \includegraphics[width=0.3\linewidth]{example-image} &
%         \includegraphics[width=0.3\linewidth]{example-image} \\
%     \end{tabular}
%     \caption{Placeholder: Effect of size and position of the spherical light source. \todo{show effect of distance, size, etc. of light source} \todo{self-shadowing effect}}
%     \label{fig:effect-light-size-position}
% \end{figure}

\input{figs/figure_ablation_cfg_weight}
% \input{figs/figure_ablation_latent_weight}
% \input{figs/figure_ablation_neg_sample}

% \begin{figure}[!ht]
%     \centering
%     \begin{tabular}{ccc}
%         \includegraphics[width=0.3\linewidth]{example-image} &
%         \includegraphics[width=0.3\linewidth]{example-image} &
%         \includegraphics[width=0.3\linewidth]{example-image} \\
%         \includegraphics[width=0.3\linewidth]{example-image} &
%         \includegraphics[width=0.3\linewidth]{example-image} &
%         \includegraphics[width=0.3\linewidth]{example-image} \\
%         Shadow overlay & Masked shading & Ours
%     \end{tabular}
%     \caption{Placeholder: Ablation study on type of shadow blending.}
%     \label{fig:ablation-shadow}
% \end{figure}

%% file: figs/figure_ablation_cfg_weight.tex
\begin{figure}[!ht]
    \centering
    \footnotesize
    
    \def\arraystretch{0.5}
    \setlength{\tabcolsep}{1pt}
    \setlength{\mywidth}{0.325\linewidth}
    
    \begin{tabular}{ccc}
        \includegraphics[width=\mywidth]{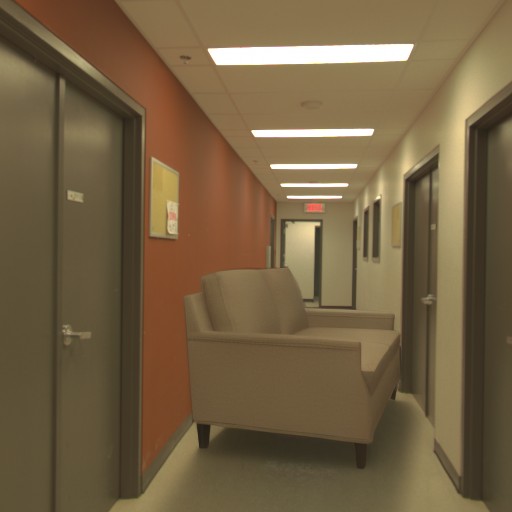} &
        \includegraphics[width=\mywidth]{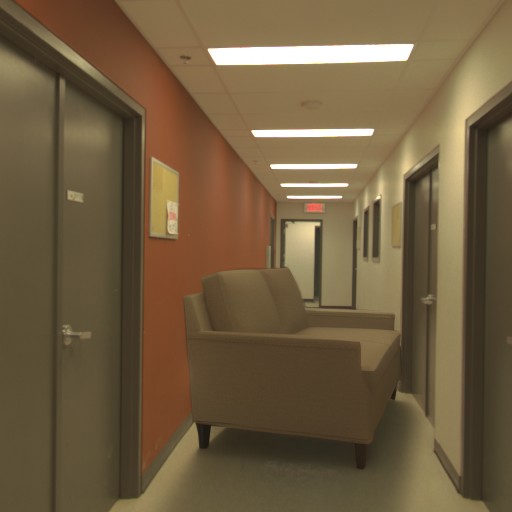} &
        \includegraphics[width=\mywidth]{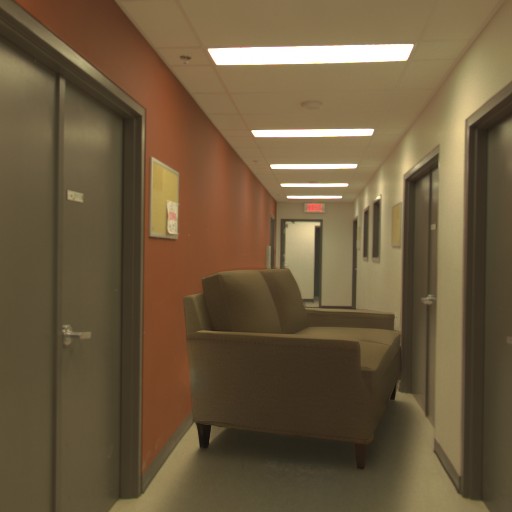} \\
        \includegraphics[width=\mywidth]{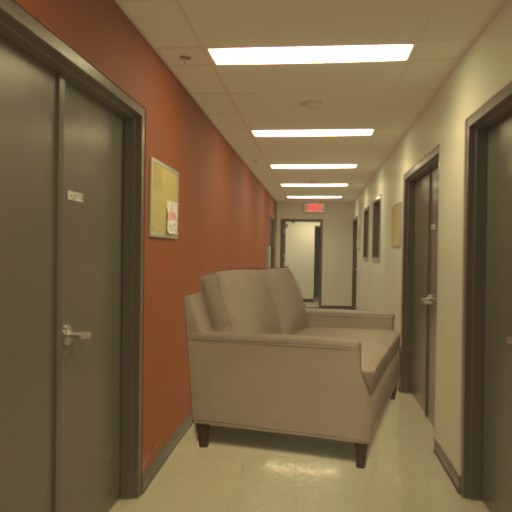} &
        \includegraphics[width=\mywidth]{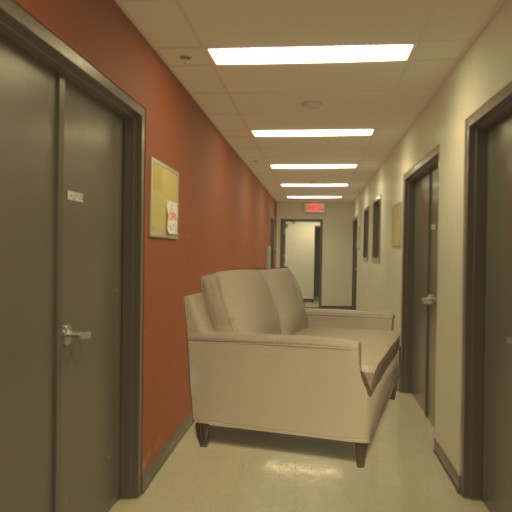} &
        \includegraphics[width=\mywidth]{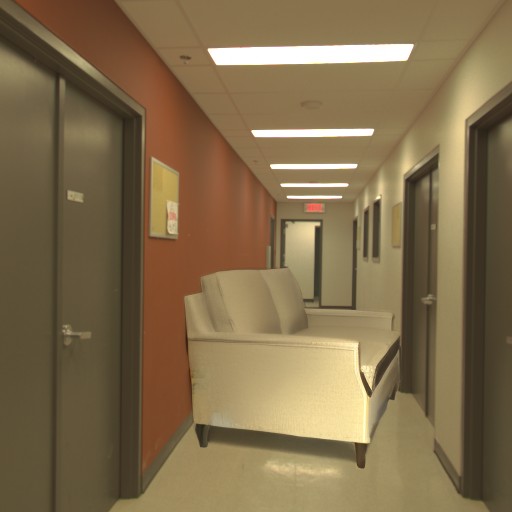} \\
        No guidance ($\gamma = 1$) & 
        $\gamma = 3$ (ours) & 
        $\gamma = 7$
    \end{tabular}
    \vspace{-5pt}
    \caption{Effect of varying the local guidance scale $\gamma$, on back (top) and front (bottom) lights. A low $\gamma$ value (left) results in minimal variation in the relighting effect, whereas a high value (right) over-amplifies the reshading from the virtual light source. We empirically use $\gamma=3$ as it offers the best balance between realism and control for indoor scene settings.}
    \label{fig:ablation-cfg-scale}
\end{figure}

%% file: sec/extensions.tex
\section{Applications}
\label{sec:extensions}

This section highlights additional capabilities derived from our framework to further demonstrate its versatility.

\begin{figure}
    \centering
    \def\arraystretch{0.25}
    \setlength{\tabcolsep}{1pt}
    \setlength{\mywidth}{0.325\linewidth}
    \begin{tabular}{ccc}
     \includegraphics[width=\mywidth]{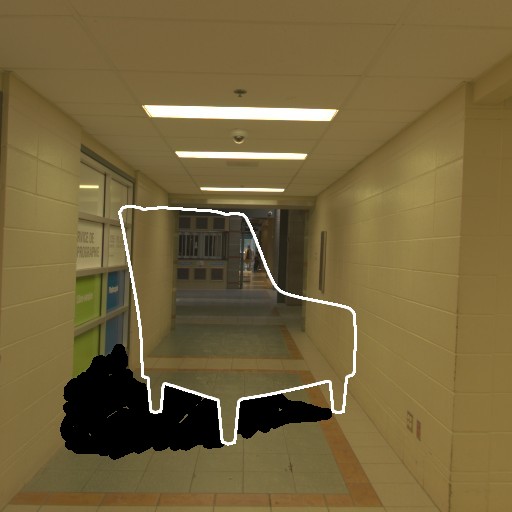} &
        \includegraphics[width=\mywidth]{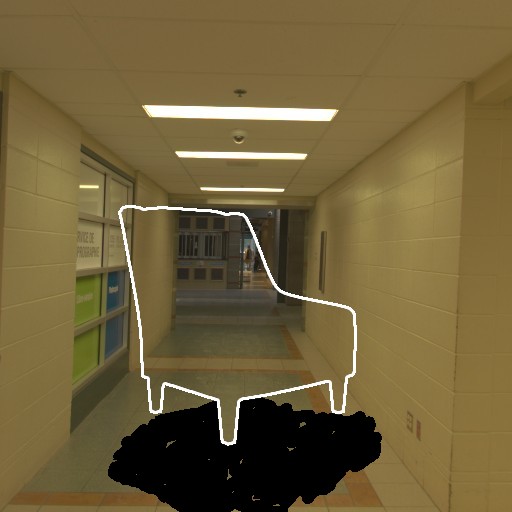} & \includegraphics[width=\mywidth]{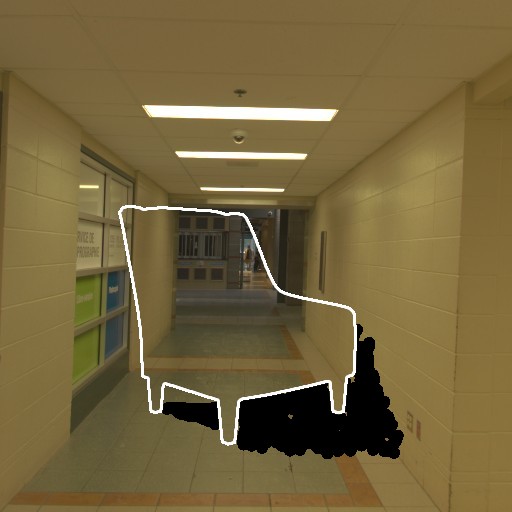} \\
     \includegraphics[width=\mywidth]{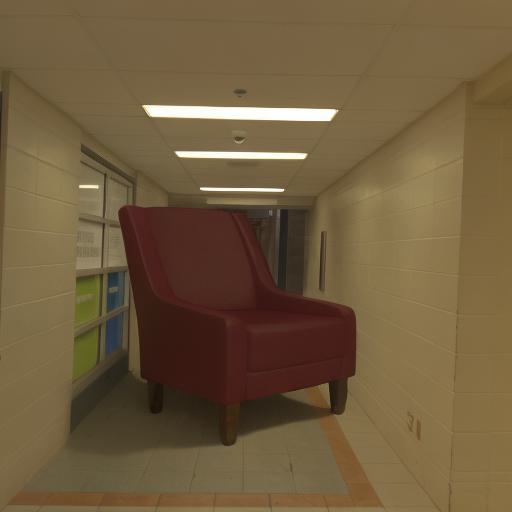} &
        \includegraphics[width=\mywidth]{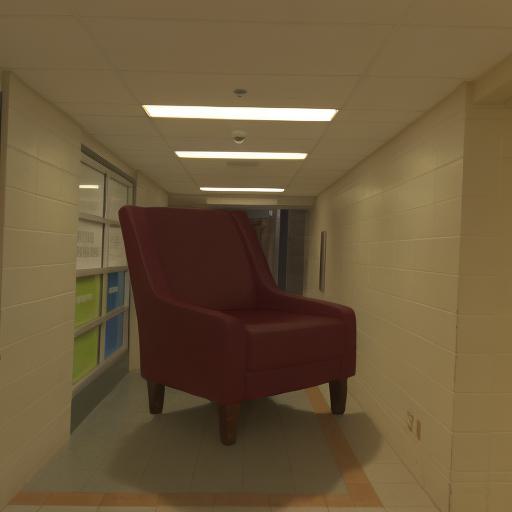} & \includegraphics[width=\mywidth]{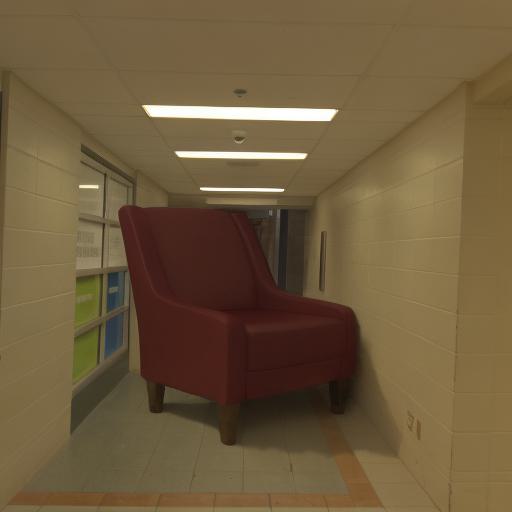} 
        
    \end{tabular}
    \vspace{-5pt}
    \caption{Rough sketches of shadows also provide a powerful lighting cue to \method. By scribbling a shadow next to the object region (top), \method is able to realistically relight the object and refine the shadow (bottom). }
    \vspace{-5pt}
    \label{fig:shadow-scribbles}
\end{figure}

\myparagraph{Scribbles as shadows.} 
Instead of relying on shadow mapping in 3D, one can simply \emph{draw} the desired shadow! \Cref{fig:shadow-scribbles} shows that \method can realistically relight the object and refine its shadow even when it is a user-drawn scribble. 

% , we used shadow generation methods that are parametrized by a point light such as shadow mapping or pixel height shadow estimator. We, however, demonstrate that using a simple user-drawn scribble for the shadow guidance can be used to locally control the shading of the object in \cref{fig:shadow-scribbles}.

% \myparagraph{Multiple light sources.}
% We can combine individual outputs from \method at different light directions (obtained with different guiding shadow maps) to simulate multiple light sources. Please refer to the supp. for examples.

\begin{figure}[t]
\footnotesize
    \setlength{\tabcolsep}{1pt}
    \begin{tabular}{cccc}
    \includegraphics[width=0.24\columnwidth]{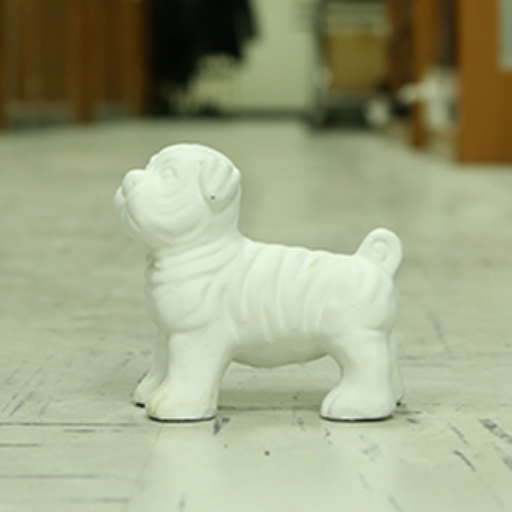} & \includegraphics[width=0.24\columnwidth]{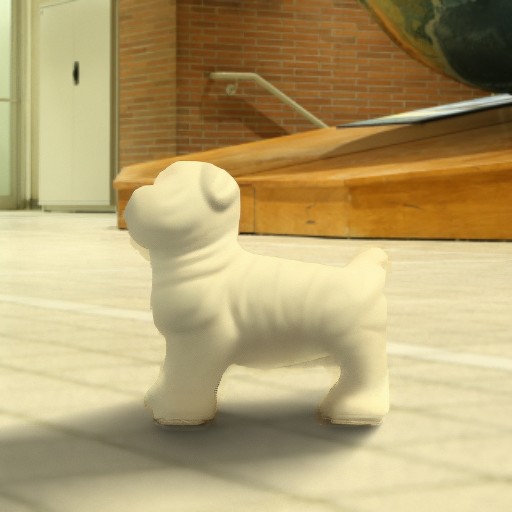}  & \includegraphics[width=0.24\columnwidth]{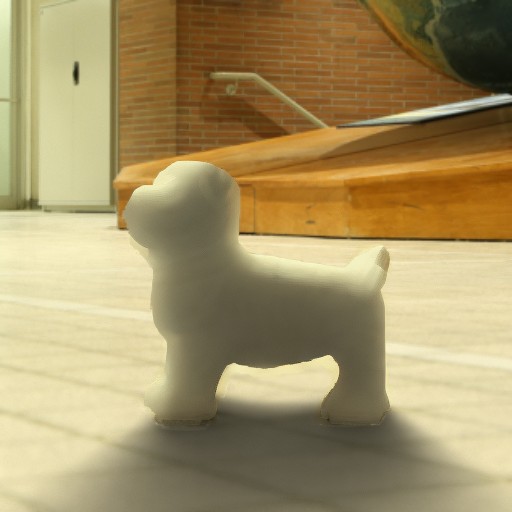} & \includegraphics[width=0.24\columnwidth]{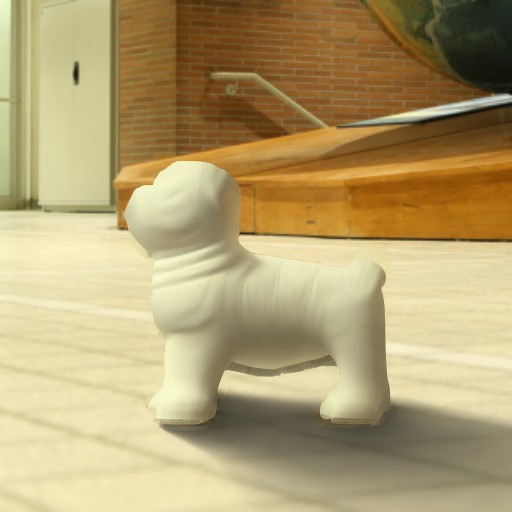} \\
    Input object & \multicolumn{3}{c}{Relit object on target background}
    \end{tabular}
    \vspace{-5pt}
    \caption{\method can also be applied to objects from 2D images using estimated intrinsics and shadows generated using \cite{tasar2024controllable}.}
    \label{fig:results-2d}
    \vspace{-5pt}
\end{figure}

\myparagraph{2D objects.}
%We demonstrate \method can be applied to real 2D objects from existing photographs by estimating their intrinsics, as in \cref{sec:eval-specifics}, and generating desired shadows with shadow generation methods such as \cite{sheng2023pixht} or, as illustrated in \cref{fig:results-2d}, with specially-designed networks~\cite{tasar2024controllable}.
\method can be applied to real 2D objects from existing photographs by estimating their intrinsics, as in \cref{sec:eval-specifics}, and generating desired shadows with shadow generation methods such as~\cite{tasar2024controllable}, as illustrated in \cref{fig:results-2d}.

\begin{figure}
\newcommand{\myoverlay}[1]{\includegraphics[width=0.32\columnwidth]{#1}\llap{\raisebox{1.2cm}{\frame{\includegraphics[width=0.22\columnwidth, trim=11cm 2cm 0 10cm, clip]{#1}}}\hspace{0.9cm}}}
    \setlength{\tabcolsep}{1pt}
    \begin{tabular}{ccc}
    \myoverlay{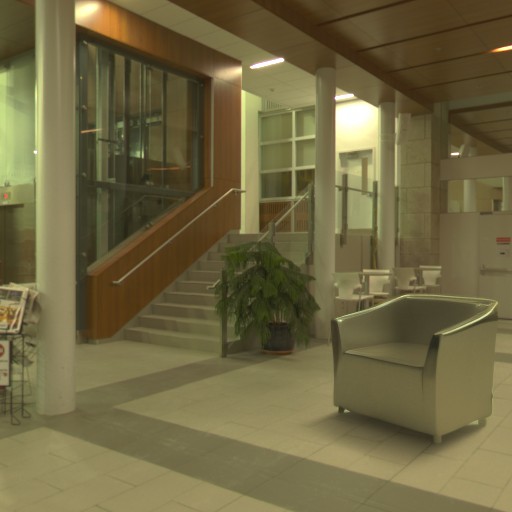} &
    \myoverlay{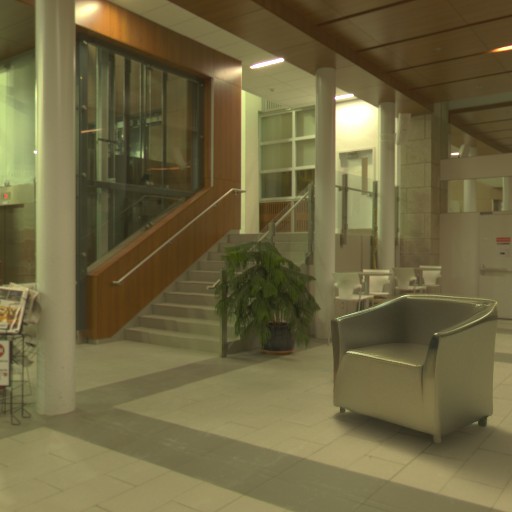}  &
    \myoverlay{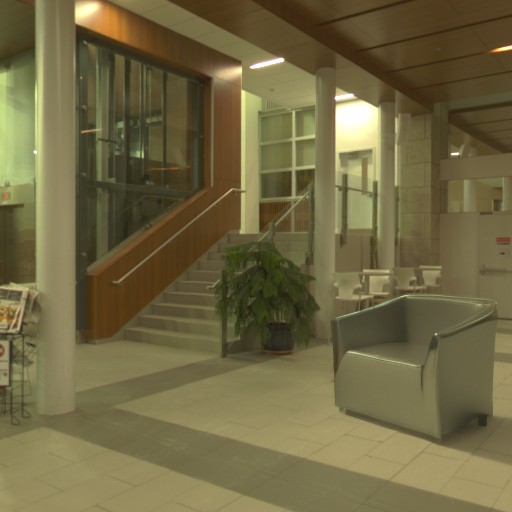} \\
    \end{tabular}
    \vspace{-5pt}
    \caption{When using a ZeroComp backbone trained on InteriorVerse~\cite{zhu2022learning}, \method generates reflections on specular objects.}
    \label{fig:metallic}
\end{figure}

\myparagraph{Reflective materials.}
When using a ZeroComp backbone pre-trained on the InteriorVerse dataset~\cite{zhu2022learning}, \method naturally generalizes to metallic objects and generates controllable specular reflections on the object, see \cref{fig:metallic}.

\myparagraph{Additional relighting results.} \method also shows other capabilities: relighting for more diverse objects and outdoor scenes, simulating multiple light sources by combining outputs with different guiding shadows, and full-image relighting. Due to space constraints, the implementation details and results are provided in the supp.

%% file: sec/discussion.tex
\section{Discussion}
\label{sec:discussion}
\begin{figure}
    \centering
    \footnotesize
    \def\arraystretch{0.5}
    \setlength{\tabcolsep}{1pt}
    \setlength{\mywidth}{0.24\linewidth}
    \begin{tabular}{cccc}
\includegraphics[width=\mywidth]{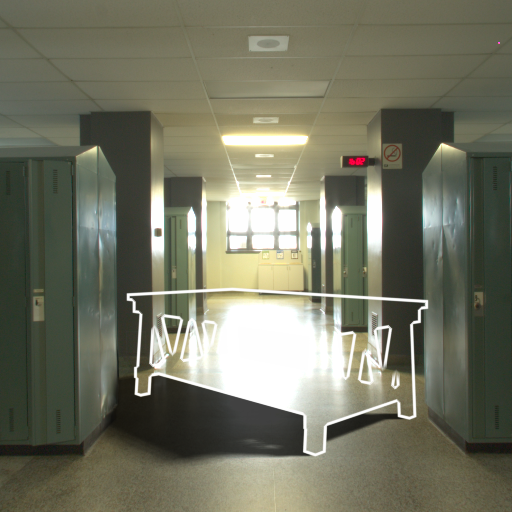} & 
\includegraphics[width=\mywidth]{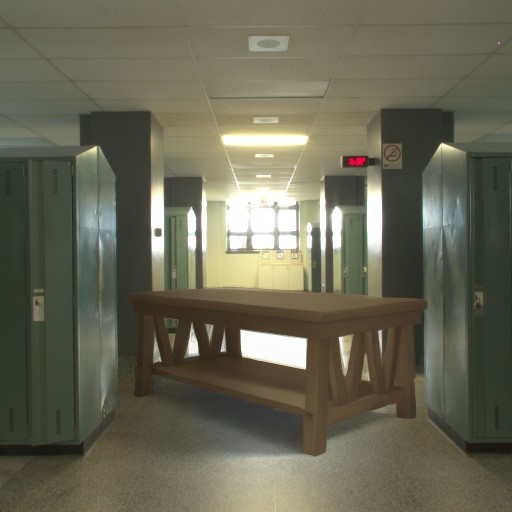} & 
\includegraphics[width=\mywidth]{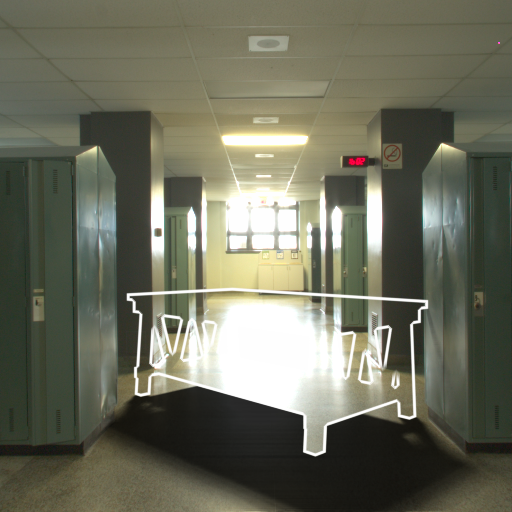} & \includegraphics[width=\mywidth]{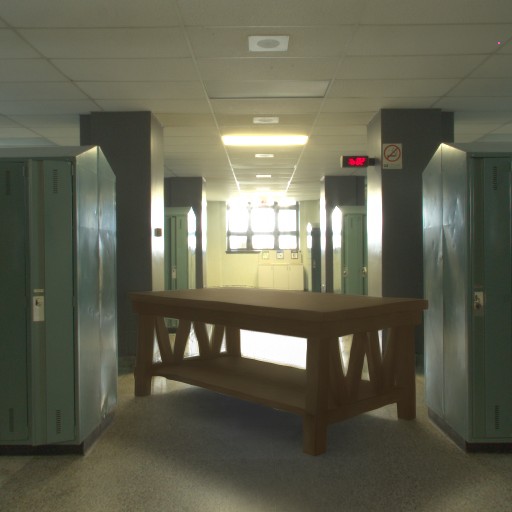} \\
 \multicolumn{2}{c}{Incorrect user-defined shadow} & 
 \multicolumn{2}{c}{Correct shadow}
    \end{tabular}
    \vspace{-5pt}
    
    \caption{User-specified (possibly inconsistent) shadows. Even when the input shadow contradicts the scene lighting (left), \method produces a visually plausible composition; the user may adjust shadow direction (right).}
    \label{fig:physically_incorrect_shadows}
\end{figure}

\myparagraph{Limitation: physically incorrect shadows.} Since \method imposes no constraints on the input shadow, users may specify a shadow direction that potentially contradicts the real lighting in the background. In this case, \method still attempts to render a visually plausible composition consistent with the given shadow, as shown in \cref{fig:physically_incorrect_shadows}.

We present \method, a method for realistically controlling the local lighting of an object through a coarse shadow, compatible with diffusion-based intrinsic image renderers without any additional training. Our results demonstrate that fine-grained control over local lighting can be achieved while attaining realistic compositions.

While our method can realistically generate relighting results from a shadow and despite showing that this can be done in multiple ways (shadow mapping, 2D generation, hand-drawn scribbles), the process of generating a shadow from scratch could prove challenging to some. A promising future direction is to enable end-to-end shadow control using parametric light models, such as point or sphere lights. Additionally, similar approaches could be explored to enable global lighting adjustments for entire images.

%Limited background relighting capabilities.

%Requires a trained diffusion neural renderer, that is, a model trained to render from intrinsics. In future work, we would like to investigate the applicability of \method to other diffusion backbones, which do not rely on intrinsics.

%No temporal consistency.

%Requires the shadow (or a negative shadow) to be visible.